\documentclass[10pt,twocolumn,letterpaper]{article}

\usepackage[pagenumbers]{cvpr}

\usepackage{url}           
\usepackage{booktabs}
\usepackage{amsfonts}      
\usepackage{nicefrac}      
\usepackage{microtype}     
\usepackage{xcolor}        
\usepackage{wrapfig} 

\usepackage{adjustbox}
\usepackage{diagbox}
\usepackage{multirow}
\usepackage{makecell}

\usepackage{times}
\usepackage{epsfig}
\usepackage{graphicx}
\usepackage{amsmath}
\usepackage{amssymb}
\usepackage{colortbl}
\usepackage{xspace}
\usepackage{tcolorbox}

\usepackage{ragged2e}
\usepackage{float}
\usepackage{bbding}
\usepackage{hhline}
\usepackage{nicematrix}
\usepackage{capt-of}
\usepackage{amssymb}  
\usepackage[accsupp]{axessibility}
\usepackage{bbding}
\usepackage[colorlinks=true, linkcolor=blue, citecolor=green]{hyperref}
\usepackage[capitalize]{cleveref}
\definecolor{cvprblue}{rgb}{0.21,0.49,0.74}

\title{Video2Act: A Dual-System Video Diffusion Policy with Robotic \\ Spatio-Motional Modeling}

\author{
Yueru Jia\textsuperscript{\rm 1,2$^{*}$}, Jiaming Liu\textsuperscript{\rm 1$^{* \dagger}$}, Shengbang Liu\textsuperscript{\rm 1$^{*}$},  Rui Zhou\textsuperscript{\rm 1}, Wanhe Yu\textsuperscript{\rm 1}, 
Yuyang Yan\textsuperscript{\rm 1},\\ Xiaowei Chi\textsuperscript{\rm 3}, Yandong Guo\textsuperscript{\rm 2}, Boxin Shi\textsuperscript{\rm 1}, Shanghang Zhang\textsuperscript{\rm 1}\textsuperscript{\Envelope}
\vspace{0.2cm}\\
\textsuperscript{\rm 1}State Key Laboratory of Multimedia Information Processing, School of Computer Science, \\ Peking University; 
\textsuperscript{\rm 2}AI$^2$Robotics; \textsuperscript{\rm 3}Hong Kong University of Science and Technology\\
$^{*}$Equal Contribution, $^{\dagger}$Project lead, \Envelope Corresponding Author \vspace{0.2cm} \\
\textbf{Project Page:} \href{https://video2act.github.io/}{https://video2act.github.io/}
}
\begin{document}
\twocolumn[
{%
\renewcommand\twocolumn[1][]{#1}
\maketitle

\begin{center}
\centering
\begin{minipage}[t]{\linewidth}
\includegraphics[width=\textwidth]{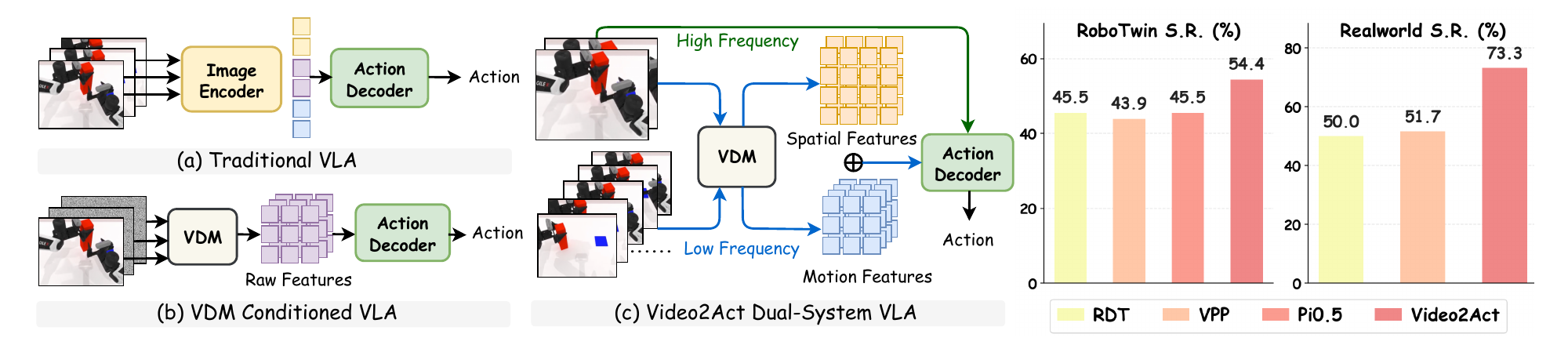}
\vspace{-0.6cm}

{\captionsetup{hypcap=false}
\captionof{figure}{\footnotesize{\textbf{Video2Act.}
Our asynchronous dual-system model leverages a perceptual video diffusion model for spatial–motion understanding and a fast action decoder for stable control, achieving state-of-the-art performance in both simulation and real-world tasks.}}}

\label{fig:teaser}
\end{minipage}
\end{center}

}]

\begin{abstract}

Robust perception and dynamics modeling are fundamental to real-world robotic policy learning. 
Recent methods employ video diffusion models (VDMs) to enhance robotic policies, improving their understanding and modeling of the physical world.
However, existing approaches do not fully exploit the coherent and physically consistent motion representations inherently encoded across frames in VDMs.
To this end, we propose Video2Act, a framework that guides robotic action learning by explicitly integrating spatial and motion-aware representations from VDMs. Instead of relying on raw VDM tokens, we extract foreground object boundaries and inter-frame motion variations that emphasize interaction-centric cues while filtering out background distractions and task-irrelevant biases. These structured representations are injected as conditioning signals into a diffusion transformer (DiT) expert, enabling it to reason about what to manipulate and how to move.
To effectively leverage these structured cues, we introduce a representation-driven dual-system architecture. In this design, the VDM operates as a slow System 2, while the DiT action expert functions as a fast System 1, jointly grounding stable, high-frequency action generation in temporally consistent motion representations.
For evaluation, Video2Act surpasses previous state-of-the-art VLA methods by 8.9\% in simulation and 21.7\% in real-world tasks in terms of average success rate, further exhibiting strong generalization capabilities.
% \keywords{Video Diffusion Models \and Vision-Language-Action Models \and Dual-System Architecture}
\end{abstract}
    
\section{Introduction}
\label{sec:intro}
% \begin{figure*}
%   \centering
%   \includegraphics[width=\textwidth]{images/new_fig/fig_01_v4.pdf}
%   \caption{\textbf{Video2Act.} 
%   Our asynchronous dual-system model leverages a perceptual video diffusion model for spatial–motion understanding and a fast action decoder for stable control, achieving state-of-the-art performance in both simulation and real-world tasks.}
% \end{figure*}

Robot learning aims to acquire action policies through interaction with the environment, leveraging dynamic visual observations to guide decision-making~\cite{levine2016end, kalashnikov2018scalable, lynch2020learning}. 
From a cognitive perspective, humans acquire manipulation skills not only by understanding the spatial structure of manipulated objects but also by abstracting transferable motion patterns across tasks~\cite{lake2017building}. 
{For example, when closing a laptop lid, humans rely on understanding the spatial relationship between the screen and the hinge, while abstracting a motion sequence that involves first grasping the screen and then closing it.
This insight suggests that effective robotic policy learning requires representations expressly designed for robust perception and dynamics modeling.}

{A straightforward approach is to concatenate visual frames extracted by a static visual encoder, but this merely allows the model to observe multiple frames without enabling it to understand the temporal and causal dependencies among them.
The recent success of video diffusion models (VDMs)~\cite{ho2022video} in capturing realistic dynamics demonstrates their strong ability to model physical environments.
Inspired by this, recent advances have incorporated representations derived from VDMs to enhance robotic policy learning, enabling a richer understanding of scene context and future evolution.
However, most existing policies~\cite{hu2024video, chi2025mind} do not explicitly explore the spatial structures and motion dynamics inherently encoded in the raw representations of VDMs, which could serve as more efficient and informative conditions for action learning.}

\begin{figure*}[t]
  \centering
  \includegraphics[width=\textwidth]{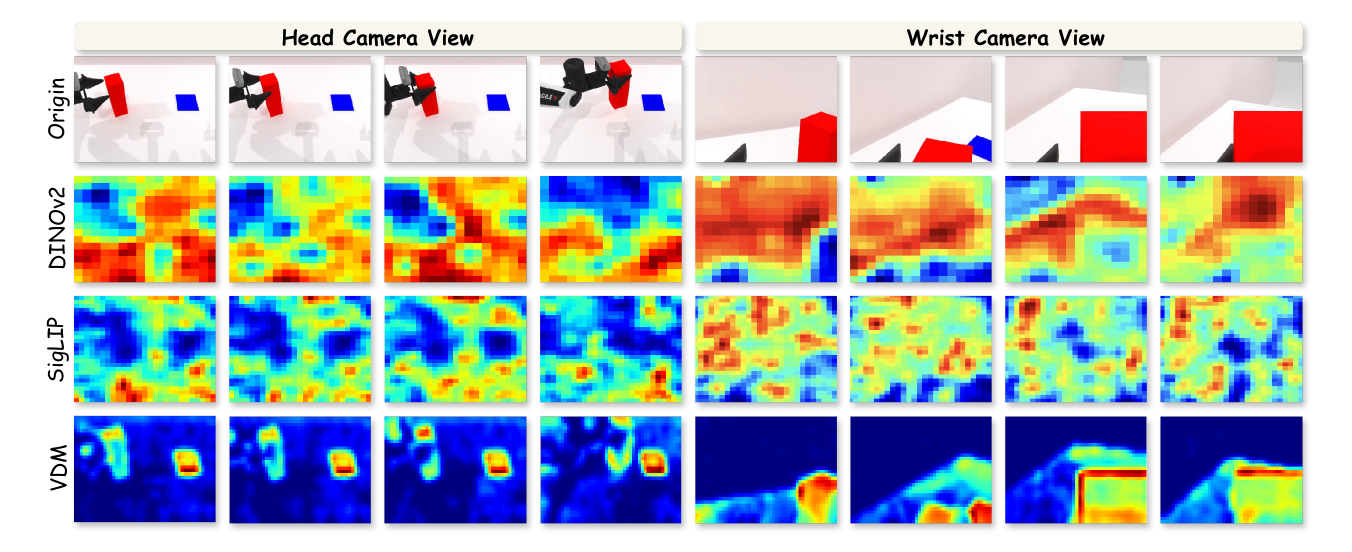}
  \vspace{-5mm}
  \caption{\textbf{Qualitative analysis of latent representations.} We visualize Grad-CAM activations for DINOv2, SigLIP and the Video Diffusion Model (VDM) during the block handover task, observed from two common robotic settings: a static third-person (Head Camera View) and a dynamic ego-centric (Wrist Camera View). The heatmaps for standard image encoders (DINOv2, SigLIP) are diffuse, unstable, and shift focus irregularly. In contrast, the VDM features consistently attend to the foreground objects, demonstrating strong spatial structure awareness even under severe ego-motion. }
  \label{fig:motivation}
\vspace{-3mm}
\end{figure*}

{To investigate the spatio-motional information encoded within VDMs, we conduct a systematic analysis of their latent features~\cite{peebles2023scalable}, which motivates our design.
Following the inversion technique applied to real videos~\cite{song2020denoising, pondaven2025video}, we extract clean feature signals from the early inversion stages to suppress noise interference.
We design two robotic validation scenarios: (1) capturing manipulation trajectories with a static third-person camera, and (2) observing manipulated objects using a dynamic wrist camera.
The inferred features from both settings are visualized with Grad-CAM~\cite{selvaraju2017grad}.
As shown in Figure~\ref{fig:motivation}, compared with static vision encoders (e.g., SigLIP~\cite{zhai2023siglip} and DINOv2~\cite{oquab2023dinov2}), VDMs demonstrate a stronger ability to capture the structural and motion consistency of foreground objects, while mitigating disturbances caused by robot and camera movement.
}

% \textcolor{red}
Building on these findings, we introduce \textbf{Video2Act}, a vision-language-action (VLA) model that explicitly integrates \textbf{spatial- and motion-aware representations} from a VDM into policy action learning. 
As shown in Figure~\ref{fig:01}, for spatial cues, we innovatively apply Spatial Filtering Operators to high-resolution images to extract salient foreground boundaries while filtering out background noise and task-irrelevant biases.
For motion cues, we leverage the Fast Fourier Transform (FFT) on long-frame sequences to highlight the motion dynamics of both the robotic arm and the manipulated objects.
Both representations are injected into the DiT action head via cross-attention conditioning, enabling the model to understand \textit{what} to manipulate and \textit{how} to move.
% Given the computational cost of the VDM, we design an asynchronous dual-system strategy.
% \textcolor{blue}{We further adopt a representation-driven asynchronous dual-system strategy to mitigate the computational cost of the VDM.} In this framework, the VDM functions as a slow perceptual module (System 2) that provides spatio-motional conditions, while the DiT action expert serves as a fast execution module (System 1), receiving real-time visual inputs and low-frequency features from System 2.
% \textcolor{blue}{Unlike prior dual-system designs, our approach explicitly leverages the temporally structured representations of the VDM, particularly through motion filters that extract temporally coherent motion representations, enabling System 1 to maintain stable control even when the perceptual module operates at a lower update frequency.} Notably, we find that by explicitly providing motion-aware conditions to System 1, it can adaptively generate stable actions when receiving varying frequency inputs.
To effectively operationalize these structured cues, we further propose a representation-driven dual-system architecture. In this framework, the VDM acts as a slow perceptual module (System 2) that extracts spatio-motional representations, while the DiT action expert serves as a fast execution module (System 1) that generates stable actions conditioned on both real-time visual observations and low-frequency features from System 2.
Unlike prior dual-system designs~\cite{figure2024helix, chenfast}, our approach leverages motion filters to extract temporally coherent representations from VDM latents. These temporally robust representations allow System 1 to maintain stable control even when System 2 operates at a lower update frequency, while also enabling Video2Act to generate adaptive actions under varying System 2 update rates.

For evaluation, we assess the multi-task performance of Video2Act on 12 RoboTwin simulation tasks~\cite{Mu_2025_CVPR, chen2025robotwin} and six real-world manipulation tasks using the ALOHA dual-arm robot~\cite{Zhao-RSS-23}. Video2Act achieves outstanding performance, outperforming previous SOTA methods by {8.9\% in simulation and 21.7\%} in real-world experiments in terms of average success rate. Moreover, it operates at a real-time action generation frequency, demonstrating both high efficiency and robust control capabilities. In summary, our contributions are as follows:

\begin{itemize}
    % \item {We systematically analyze VDM representations in robotic settings, revealing that they capture stable structural and motion-consistent features that are robust to robot arm and wrist-camera dynamics.}

    \item We conduct a systematic motivation analysis of VDM representations in robotic settings, revealing that they capture stable structural and motion-consistent features that are robust to robot arm and wrist-camera dynamics.
    
    \item {We propose Video2Act, which explicitly incorporates spatio-motional representations into VLA model learning via Spatial Filtering Operators and FFT, enabling the model to reason about what to manipulate and how to move.}
    % \item {We develop an asynchronous dual-system strategy, where the VDM acts as a slow perceptual system and the DiT head as a fast execution system, enabling adaptive and stable action generation under high-frequency inputs.}
    % \item We develop a representation-driven asynchronous dual-system strategy, where the VDM acts as a slow spatio-temporal modeling system system and the DiT head as a fast execution system, enabling stable and adaptive action generation under high-frequency control.
    \item We introduce a representation-driven dual-system architecture, where a VDM-based System 2 extracts temporally coherent spatio-motional representations to guide a fast DiT-based System 1 for stable and adaptive action generation.
\end{itemize}

\section{Related Work}

% 1. Video diffusion models for robotics：vdm的发展越来越迅速，在robotics中也有广泛的应用。目前的有一些方法显示的预测未来状态，xxx；在之后有一些工作开始提取feature等等。而video diffusion models本身所具有的一些信息并没有被很好的discuss.
% - Video diffusion models (VDMs) have recently shown remarkable ability in modeling complex temporal and physical dynamics across video sequences.
% - Our work explicitly extracts and exploits these motion-consistent cues to guide robotic action generation.
% 2. VLA: 概括一下vla的现状，然后说目前一些方法使用了双系统.
%   1. Vision-Language-Action (VLA) Models
%   2. Recent works further explore dual-system architectures that decouple slow reasoning from fast action execution, inspired by xxx. Such designs enhance sample efficiency and real-time adaptability but often rely on static perception inputs.
%   3. In contrast, our framework introduces a spatial and motion-aware asynchronous dual system, where the slow system (VDM) captures global temporal dynamics, and the fast system (DiT-based action head) performs adaptive control conditioned on motion-consistent cues.

\subsection{Video Diffusion Models for Robotics}
% Recently, researchers find the strong motion modeling ability of video generation, and its potential in robotics planning. Earlier works like VideoGPT\cite{}, UniSim\cite{}, using robot video data to training the Transformer or diffusion prediction model, and transform motion into robot action through inverse-dynamic-model\cite{}. To better implement the action modeling, AVDC\cite{} using flow to caption the motion, Anygras\cite{} further improve the projection quality, Robodreamer optimize the generation quality, and Manipulate3D improve through trustwise replanning. More recent works using VDM as a backbone for VLA, UVA, MinD, and more works using VDM to extract feature to enhance the traditional VLA, like dreamVLA, UniVLA, etc. However, IDM face significant limiation of VDM generation efficiency, while VLA based using inplicity features, ignore the generation models's inner generalizablity. Our work explicitly extracts and exploits these motion-consistent cues to guide robotic action generation.
Recent progress in video diffusion models (VDMs) has significantly advanced the ability to generate temporally consistent video sequences~\cite{ho2022video}. VDMs learn to model long-horizon dynamics through iterative denoising in the latent space, enabling strong representations of object motion, scene transitions, and causal dependencies across time. These capabilities have led to growing interest in using VDMs for downstream robotic perception and control~\cite{finn2016unsupervised, liao2025genie, yang2023learning, hu2023look, zhou2024robodreamer, chi2025wow}. Early works use VDMs to predict future videos and employ inverse dynamics models to infer actions~\cite{ajay2023compositional, ko2023learning, luo2024grounding, bruce2024genie}, but these approaches often suffer from accumulated errors and high inference latency. More recent work explores leveraging VDMs in an implicit manner; for example, VPP~\cite{hu2024video} learns to extract intermediate feature maps from VDMs and inject them into policy networks to end-to-end learn action policies. Despite these advances, existing approaches typically directly use raw VDM features without fully understanding or disentangling the spatial structures and motion dynamics encoded within them~\cite{liao2025genie, li2025unified, chi2025mind}. Such structural and temporal information remains underutilized, even though it provides strong priors for action learning. Our work addresses this gap by explicitly extracting and refining spatial- and motion-aware representations, and integrating them as conditions for diffusion-based action policy learning.

% Recent advancements in VLA modeling have enabled robots to follow natural language instructions to perform manipulation tasks ~\cite{liu2025hybridvla, black2024pi_0}. Approaches such as diffusion policies ~\cite{chi2023diffusion} and transformer-based policies ~\cite{kim2024openvlaopensourcevisionlanguageactionmodel, wen2025tinyvla, gao2025adaworld} have shown promise in learning complex visuomotor behaviors. Noticing the frame-rate limitation of the large-scale backbone, dual-system design such as \cite{}
\subsection{Vision-Language-Action (VLA) Models}

Recent progress in developing robust and generalizable manipulation policies has led to the emergence of Vision–Language–Action (VLA) models, which integrate visual observations, natural language instructions, and policy learning into a unified framework~\cite{rt22023arxiv, li2024manipllm, kim2024openvla, Black2024pi0, pertsch2025fast, wen2024diffusion, wen2024tinyvla}. Most VLAs employ image encoders to extract visual representations that are fused with language embeddings to guide an action decoder~\cite{octo_2023, liu2025hybridvla, intelligence2025pi05visionlanguageactionmodelopenworld}, which limits their ability to capture long-horizon temporal dynamics and detailed spatial structure in complex environments. To reduce the computational overhead of large vision–language models, several recent approaches introduce a dual-system decomposition, separating high-level reasoning from low-level action generation~\cite{han2024dualprocessvlaefficient, bu2025synergisticgeneralizedefficientdualsystem, bjorck2025gr00t, zhang2024hirt, figure2024helix, chenfast}. In these frameworks, System 2 performs semantic or task-level processing to produce a latent representation that conditions a diffusion-based System 1 action policy. In our work, the VDM is also computationally expensive, and its strong temporal modeling capability naturally fits the role of System 2 within such a dual-system design.
\section{Method}
% We begin by analyzing the feature representations of the video DiT under two common robotic observation settings—a static third-person camera view and a dynamic wrist-mounted camera view (Section~\ref{subsec:dit-analysis}). The feature analysis in Section~\ref{subsubsec:feature-analysis} reveals that the video DiT activations consistently emphasize foreground objects and remain temporally stable, indicating that the model inherently encodes spatial structures and motion dynamics relevant to robotic manipulation.
% \textcolor{blue}{TODO: action distribution}
% 补充一个action分布
To explore the inherent spatio-motional information encoded within video diffusion models (VDM), we first conduct a qualitative analysis of their latent feature representations under two common robotic observation settings: a static third-person camera view and a dynamic wrist-mounted camera view (Section~\ref{subsec:dit-analysis}). The results show that VDM activations consistently emphasize foreground objects and exhibit high temporal stability, revealing strong spatial structure and motion-aware characteristics. Motivated by this observation, we introduce the Video2Act model in Section~\ref{subsec:Video2Act}, a vision-language-action (VLA) framework that explicitly incorporates spatial- and motion-aware representations extracted from the VDM. To support real-time control, we design an asynchronous dual-system scheme, presented in Section~\ref{subsec:dual-system}, where the VDM operates as System 2, providing perceptual updates, while a DiT-based action head serves as System 1, producing high-frequency action generation.

% To support real-time control, we further design an asynchronous dual-system scheme, presented in Section~\ref{subsec:dual-system}, where the VDM operates as System 2, providing low-frequency perceptual updates, while a DiT-based action head serves as System 1, producing high-frequency action generation.
% Motivated by this observation, we introduce Video2Act in Section~\ref{subsec:Video2Act}, a 
% vision-language-action (VLA) framework that explicitly incorporates spatial- 
% and motion-aware representations extracted from the video DiT, which are 
% described in detail in Section~\ref{subsec:extraction}. To support real-time control, we further 
% adopt an asynchronous dual-frequency execution scheme described in Section~\ref{subsec:dual-system}, 
% where the video DiT provides low-frequency perceptual updates as System 2, 
% while a DiT-based action head produces high-frequency control outputs as 
% System 1.

\subsection{Motivation}
\label{subsec:dit-analysis}
% \subsubsection{Feature Analysis in Robotic Settings}
% \label{subsubsec:feature-analysis}
Existing VLA methods~\cite{Black2024pi0, kim2024openvla, liu2025hybridvla, liu2024rdt, lin2025onetwovla} rely on static image encoders of pretrained VLMs to extract visual features. However, such image encoders lack the temporal perception and dynamics modeling capabilities required for robotic manipulation.
Recently, motivated by the strong ability of video diffusion models (VDMs) to capture realistic dynamics in the physical world, several VLA approaches~\cite{hu2024video, chi2025mind} have begun to directly incorporate raw representations derived from VDMs to improve robotic policy learning.

However, why and how the latent features of VDMs benefit robotic manipulation has not been explicitly explored.
Therefore, we conduct an intuitive qualitative analysis of VDM latent representations under two common robotic observation settings: (1) a static scenario with a third-person camera and (2) a dynamic scenario with a wrist camera.
As shown in Figure~\ref{fig:motivation}, we visualize the Grad-CAM activations~\cite{selvaraju2017grad} of two widely used image encoders in VLA (SigLIP~\cite{zhai2023siglip} and DINOv2~\cite{oquab2023dinov2}), and compare them with those from the VDM~\cite{kong2024hunyuanvideo}. 

We use the \textit{block handover} task from RoboTwin~\cite{Mu_2025_CVPR} as the input sequence for all models to generate class-discriminative localization maps by weighting feature activations with their output gradients.
The highlighted regions indicate spatial locations that contribute most to the model’s feature response, providing an attention-like visualization of where the model focuses.
Under the static third-person view, image encoders exhibit unstable and inconsistent attention across consecutive frames, with their focus regions shifting irregularly as the manipulation progresses.

In contrast, VDM features consistently attend to the foreground object over time. Even as the block is lifted by the gripper, the highlighted regions remain stably aligned with the manipulated object, demonstrating strong spatial structure representation even under dynamic scenarios.
This consistency also persists under the dynamic wrist camera setting, where ego-motion is severe, yet the attention of the VDM remains firmly anchored to the manipulated object.

% \vspace{-2mm}
\subsection{Video2Act Framework}
\label{subsec:Video2Act}
Building on these findings, we present Video2Act, a vision-language-action (VLA) framework that explicitly integrates spatial- and motion-aware representations from a video diffusion transformer into policy learning. 

\noindent\textbf{Problem Definition.}
Following recent advances in VLA model imitation learning~\cite{kim2024openvla, liu2024rdt},
our objective is to learn a conditional policy that generates temporally coherent action trajectories conditioned on multimodal observations.
Given the observations $o_t = (I_{t-T:t}, s_t)$ and the language instruction $l$, where $I$ denotes the image sequence and $s$ represents the robot state,
the policy $\pi_{\theta}$ is trained to maximize the log-likelihood of demonstrated future actions:
$
\max_\theta \;
\mathbb{E}_{(o_t,\,l,\,a_{t+1:t+H}) \sim \mathcal{D}}
\big[\log \pi_\theta(a_{t+1:t+H} \mid o_t, l)\big].
\label{eq:policy_obj}$
The predicted action chunk $\mathbf{a}_{t+1:t+H}$ consists of joint position commands for robot control.

\begin{figure*}[t]
  \centering
  \includegraphics[width=\textwidth]{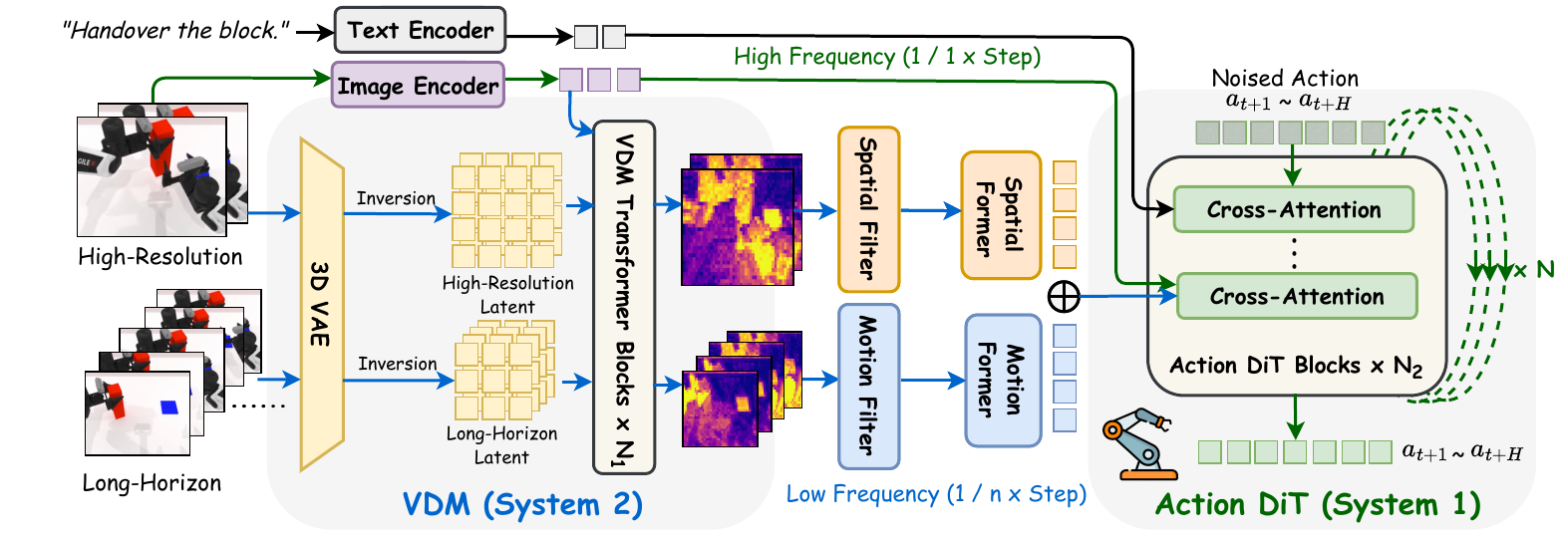}
    \vspace{-3mm}
  \caption{\textbf{Video2Act Framework.} Video2Act employs an asynchronous dual-system framework consisting of a slow perceptual VDM (System 2) and a fast action head (System 1). System 2 extracts refined spatial and motion representations from two image inputs: high-resolution images for spatial filtering via Sobel operators and long-horizon sequences for motion extraction via FFT. These low-frequency spatio-motional features serve as conditioning inputs to System 1, which simultaneously receives high-frequency image tokens. Through cross-attention conditioning, these asynchronously updated signals are effectively fused, enabling robust and real-time action generation.
  % This design enables Video2Act to achieve stable and adaptive vision-language-action control under asynchronous perception and execution.
  }
  % \vspace{-3mm}
  \label{fig:02}
\end{figure*}

\noindent\textbf{Model Architecture.}
We show the Video2Act architecture in Figure~\ref{fig:02}.
Specifically, we employ SigLIP-ViT-L/14~\cite{zhai2023siglip} as the image encoder to extract visual tokens, and a text encoder~\cite{raffel2020exploring} to produce instruction embeddings.
For System 2, we adopt the transformer-based Video Diffusion Model Hunyuan~\cite{kong2024hunyuanvideo}, which includes a pretrained 3D VAE to encode images into latent space and consists of 60 transformer blocks, from which we use the first 25 blocks to extract features.
% The layer-normalized spatial and motion features are compressed via two lightweight Q-former to reduce token redundancy while preserving global consistency.
The spatial and motion features are extracted using our proposed spatial and motion filters, and are then compressed by two lightweight Q-formers~\cite{kim2024towards} to reduce token redundancy while preserving global consistency.
For System 1, we employ a 1B-parameter diffusion transformer. The image tokens, text tokens, and VDM tokens are integrated into action DiT through cross-attention.

% The overall architecture is illustrated in Figure~\ref{fig:02}.

% \subsubsection{Overview}

% \begin{equation}
% \max_\theta \;
% \mathbb{E}_{(o_t,\,l,\,a_{t+1:t+H}) \sim \mathcal{D}}
% \big[\log \pi_\theta(a_{t+1:t+H} \mid o_t, l)\big].
% \label{eq:policy_obj}
% \end{equation}

% Further architectural details and ablation studies are provided in Appendix X.

\subsection{Spatio-Motional Representation Extraction}
\label{subsec:extraction}
% To obtain clean feature signals, we adopt an inversion-based feature extraction strategy inspired by prior diffusion inversion works~\cite{song2020denoising, pondaven2025video}. 
% Given an RGB observation sequence $\{I_t\}_{t=1}^{T}$, 
% we encode it into the latent space of the video diffusion transformer $\mathbf{V}_{\theta}$ 
% and extract features from the \textit{beginning of the inversion trajectory} ($t_{\text{diff}}{=}0$), where they preserve the original spatial structure and minimize denoising artifacts:
To obtain clean feature signals while maintaining efficiency, we adopt an inversion-based feature extraction strategy.
Following prior diffusion inversion works~\cite{song2020denoising, pondaven2025video},
given an RGB observation sequence $\{I_t\}_{t=1}^{T}$, 
we encode it into the latent space of the video diffusion transformer $\mathbf{V}_{\theta}$ 
and extract features from the \textit{beginning of the inversion trajectory} (i.e., the denoising step at $t_{\text{diff}}{=}0$), where the features preserve the original representation and minimize denoising artifacts:
\begin{equation}
    F_{t-T:t}^{} = \mathbf{V}_{\theta}(I_{t-T:t}; t_{\text{diff}}{=}0).
\end{equation}

To capture both fine-grained spatial details and long-horizon motion dynamics, 
we decompose the input into two complementary video streams processed independently by the VDM $\mathbf{V}_{\theta}$:
\begin{equation}
    F^{H}_{t-T_s:t} = \mathbf{V}_{\theta}(I^{H}_{t-T_s:t}), 
    \quad
    F^{L}_{t-T_l:t} = \mathbf{V}_{\theta}(I^{L}_{t-T_l:t}),
\end{equation}
where $H$ denotes the high-resolution input $I^{H}_{t-T_s:t} \in \mathbb{R}^{512 \times 768 \times 3}$  with the short-horizon window ($T_s = 2$), 
and $L$ denotes the normal-resolution input $I^{L}_{t-T_l:t} \in \mathbb{R}^{ 256 \times 256 \times 3}$ with the long-horizon window ($T_l = 16$), and both feature branches are extracted from the inversion step. 
% \textcolor{blue}{Ablations on $T_s$, $T_l$, and input resolution are reported in Appendix~B.3.}
Due to space limitations, we provide ablations on $T_s$, $T_l$, and the input resolution in Appendix~B.

% where $H$ and $L$ denote the \textit{high-resolution short-horizon} 
% and \textit{low-resolution long-horizon} feature branches, respectively, 
% both extracted from the inversion step at $t_{\text{diff}}{=}0$.

% As analyzed in Section~\ref{subsec:dit-analysis}, the features extracted from the video DiT 
% exhibit strong \textit{structural consistency} and \textit{temporal coherence} across frames, 
% indicating that the model inherently encodes both spatial structure and motion dynamics. 
% Building on this observation, we adopt a simple yet effective strategy to make these implicit cues explicit by applying non-learnable, channel-wise operators in the high-dimensional feature space to extract structure- and motion-aware representations. 

Building on the observations in Section~\ref{subsec:dit-analysis}, we adopt a simple yet effective strategy to make these implicit cues explicit by applying non-learnable, channel-wise operators in the high-dimensional feature space to separately extract structure-aware and motion-aware representations.
As shown in Figure~\ref{fig:02}, we apply a Sobel-based Spatial Filtering Operator to extract structural boundaries for each frame, and a frequency-domain Fast Fourier Transform (FFT) to capture coherent motion dynamics across adjacent frames.
Formally, given the two types of features 
$F^{H}_{t-T_s:t} \!\in\!\mathbb{R}^{32\times 48\times 3072}$ 
and $F^{L}_{t-T_l:t}\!\in\!\mathbb{R}^{16 \times 16\times 3072}$,
we define:
\begin{equation}
    S_t = \text{Sobel}(F^{H}_{t-T_s:t}), 
    \qquad 
    M_t = \mathcal{F}_{\text{FFT}}(F^{L}_{t-T_l:t}).
\end{equation}

\noindent\textbf{Spatial Structure Representation (Sobel).}
For each channel $c$, we compute spatial gradients over a short temporal window 
$F^{H,(c)}_{t-T_s:t}$ using the standard $3{\times}3$ Sobel kernels~\cite{sobel1968, chang2023multi}:
\[
S_x=\begin{bmatrix}-1&0&1\\-2&0&2\\-1&0&1\end{bmatrix}, \quad
S_y=\begin{bmatrix}-1&-2&-1\\0&0&0\\1&2&1\end{bmatrix}.
\]
The horizontal and vertical gradients are obtained via channel-wise convolution for each frame:
\begin{equation}
\begin{aligned}
\begin{aligned}
G_x^{(c)} = S_x * F_{t-T_s:t}^{H,(c)}, && G_y^{(c)} = S_y * F_{t-T_s:t}^{H,(c)}
\end{aligned}
\end{aligned}
\label{eq:sobel_grad}
\end{equation}

and the gradient magnitude is computed as
$S_t^{(c)} = \sqrt{(G_x^{(c)})^2 + (G_y^{(c)})^2}.$
% \begin{equation}
% \begin{aligned}
% S_t^{(c)} = \sqrt{(G_x^{(c)})^2 + (G_y^{(c)})^2}.
% \end{aligned}
% \end{equation}

The resulting representation 
$S_t = \{S_t^{(1)},\dots,S_t^{(C)}\}$ 
is obtained by applying the filter independently across channels in the high-dimensional feature space, enhancing structure-aware consistency across adjacent frames.
Empirical comparisons of different spatial filter choices are provided in Appendix~B.

\noindent\textbf{Motion Representation (FFT).}
For each spatial location $(i,j)$ and channel $c$, 
we apply a Discrete Fourier Transform~\cite{cooley1965algorithm} along the temporal axis:
\begin{equation}
\widehat{F}^{(c)}_{k}(i,j)
= \sum_{\tau=0}^{T_l-1}
  F^{L,(c)}_{t-\tau}(i,j)\,
  e^{-\,\mathrm{i}\,2\pi k\tau/T_l},
\end{equation}
where $k = 0, \dots, T_l{-}1$. We then apply a frequency mask $\mathcal{B}$, corresponding to the high-pass filter in implementation, 
and reconstruct the filtered sequence via an
inverse DFT:
\begin{equation}
M_{t-\tau}^{(c)}(i,j) =
\mathrm{Re}\!\left(
\frac{1}{T_l}
\sum_{k\in\mathcal{B}} \widehat{F}^{(c)}_{k}(i,j)\,
e^{\,\mathrm{i}\,2\pi k\tau/T_l}
\right),
\label{eq:band_inverse}
\end{equation}
where $\tau = 0, \dots, T_l{-}1.$ This operation suppresses low-frequency background components 
and highlights coherent motion patterns over time. 
The resulting tensors $S_t$ and $M_t$ explicitly encode \textit{spatial} and \textit{motion} cues 
for the action generation head.

% Table 1 in Experiment

\subsection{Representation-Driven Asynchronous Dual-System Scheme}
\label{subsec:dual-system}

Robots require efficient control to perform closed-loop manipulation tasks, yet incorporating a VDM introduces substantial computational cost.
Unlike prior dual-system designs that simply adopt heterogeneous input frequencies~\cite{figure2024helix, zhang2024hirt, chenfast}, we propose a representation-driven dual-system architecture that more effectively leverages the temporal structure of VDM representations.
Based on our key insight that spatio-motional features capture temporally consistent dynamics beyond instantaneous visual observations, these representations can be reused across time, enabling a reduced VDM inference frequency without sacrificing action generation stability.
Therefore, we propose an asynchronous dual-system paradigm within the Video2Act VLA model, where the VDM serves as System 2, a slow perceptual module that provides spatially and temporally rich representations, while a diffusion-transformer (DiT) head serves as System 1, a fast execution module that generates precise actions.
The temporally persistent semantics of System 2 allow the conditioning signals to remain informative across multiple control steps, providing the foundation for asynchronous execution in System 1.

\noindent\textbf{Asynchronous Frequency.}
Since System 2 (VDM) is a large-scale pretrained model, it operates at a low frequency to conduct high-level contextual reasoning and extract spatio-motional features. As shown in Figure~\ref{fig:02}, the feature outputs of System 2 serve as latent conditions that temporally guide System 1’s action generation over the subsequent $H$ time steps. In contrast, System 1 focuses on real-time action execution. At each time step, it uses the most recent observation to produce an action, while being conditioned on the periodically updated feature outputs from System 2. This design resembles intuitive, reactive control, positioning System 1 as a high-frequency action generation module.
Therefore, System 1 leverages the long-horizon spatio-motional representations provided by System 2 to perform fine-grained action prediction at a higher temporal resolution. Empirically (Section~\ref{exp:abaltion_ratio}), with an operating frequency ratio of $1:n$, control accuracy remains stable while inference speed improves substantially. {Further ablations indicate that this robustness stems from the temporally consistent spatio-motional representations introduced in our method.}

\noindent\textbf{Training Objective.}
System 1 (the DiT head) is trained under a conditional diffusion objective.
At each diffusion step, Gaussian noise $\epsilon$ is added to the action sequence~\cite{chi2024diffusionpolicy}, and System 1 $\mathbf{D}_{\theta}$ predicts noise conditioned on high-frequency updated image features ($F_I$), low-frequency updated VDM features ($F_{\mathrm{VDM}} = \mathrm{Cat}(S_t, M_t)$), and textual features ($F_l$), where $\tau_n$ denotes the diffusion timestep embedding: $\epsilon_{\theta} = \mathbf{D}_{\theta}\!\left(\tilde{a}_{t:t+H},\, \tau_n \mid F_I,\, F_{\mathrm{VDM}},\, F_l\right)$. These compressed tokens are injected into the DiT blocks of $\mathbf{D}_{\theta}$ via cross-attention layers. The denoising objective is defined as $\mathcal{L}_{\text{diff}} = \mathbb{E}_{a_{t+1:t+H},\,\epsilon \sim \mathcal{N}(0,1)} \big[\|\epsilon_{\theta} - \epsilon\|_2^2\big]$.

% These compressed tokens are injected into the DiT blocks of $\mathbf{D}_{\theta}$ via cross-attention layers. The denoising objective is defined as 
% \begin{equation}
% \mathcal{L}_{\text{diff}} =
% \mathbb{E}_{a_{t+1:t+H},\,\epsilon \sim \mathcal{N}(0,1)}
% \big[\|\epsilon_{\theta} - \epsilon\|_2^2\big].
% \label{eq:diff_loss}
% \end{equation}

% $\mathcal{L}_{\text{diff}} = \mathbb{E}_{a_{t+1:t+H},\,\epsilon \sim \mathcal{N}(0,1)} \big[\|\epsilon_{\theta} - \epsilon\|_2^2\big]$.

% System 1 (the DiT head) is trained under a conditional diffusion objective.
% At each diffusion step, Gaussian noise $\epsilon$ is added to the action sequence~\cite{chi2024diffusionpolicy}, and System 1 $\mathbf{D}_{\theta}$ predicts noise conditioned on high-frequency updated image features ($F_I$), low-frequency updated VDM features ($F_{\mathrm{VDM}} = \mathrm{Cat}(S_t, M_t)$), and textual features ($F_l$), where $\tau_n$ denotes the diffusion timestep embedding:
% \begin{equation}
% \epsilon_{\theta} =
% \mathbf{D}_{\theta}\!\left(
% \tilde{a}_{t+1:t+H},\, \tau_n \mid F_I,\, F_{\mathrm{VDM}},\, F_l
% \right).
% \label{eq:denoise}
% \end{equation}
% These compressed tokens are injected into the DiT blocks of $\mathbf{D}_{\theta}$ via cross-attention layers.
% The denoising objective is defined as
% \begin{equation}
% \mathcal{L}_{\text{diff}} =
% \mathbb{E}_{a_{t+1:t+H},\,\epsilon \sim \mathcal{N}(0,1)}
% \big[\|\epsilon_{\theta} - \epsilon\|_2^2\big].
% \label{eq:diff_loss}
% \end{equation}

\begin{table*}[t!]
    \centering
    \caption{
    \textbf{Simulation experiment results across RoboTwin 1.0 and 2.0 manipulation tasks.} The bold numbers indicate the highest score in each column.
    }
    \vspace{-2mm}
    \resizebox{\textwidth}{!}{%
      \setlength{\tabcolsep}{3pt} 
      \setlength{\extrarowheight}{1.5pt}
      \begin{NiceTabular}{l|cccccc|c|cccccc|c}
        \hline 
        \rowcolor{red!10!blue!10} 
        & \multicolumn{7}{c|}{\textbf{RoboTwin 1.0}} & \multicolumn{7}{c}{\textbf{RoboTwin 2.0}} \\
        \cline{2-15}
        \rowcolor{red!10!blue!10} 
        & \makecell{block \\ handover} & \makecell{container \\ place} & \makecell{cup \\ place} & \makecell{bottles \\ easy} & \makecell{bottles \\ hard} & \makecell{pick \\ apple} & \textbf{Mean} & \makecell{beat \\ block} & \makecell{click \\ bell} & \makecell{hanging \\ mug} & \makecell{place \\ object} & \makecell{place \\ shoe} & \makecell{turn \\ switch} & \textbf{Mean} \\
        \hline

        DP & 3.7 & 10.0 & 11.7 & 76.3 & 40.0 & 13.0 & $25.8 \pm 0.2$ & 44.0 & 50.7 & 3.3 & 32.0 & 25.3 & 31.7 & 31.2 $\pm$ 0.9 \\
        ACT & 13.7 & 15.0 & 23.0 & 19.3 & 19.7 & 14.3 & $17.5 \pm 2.2$ & 52.7 & 55.7 & 2.7 & 13.7 & 3.7 & 3.3 & 21.9 $\pm$ 1.1 \\
        RDT & 95.0 & 11.7 & 42.3 & 78.0 & 32.0 & 10.7 & $44.9 \pm 0.5$ & 68.7 & 72.3 & 24.3 & 41.3 & 36.3 & 33.0 & 46.0 $\pm$ 0.9 \\
        $\pi_0$ & 52.7 & 25.3 & 37.7 & 93.3 & 36.3 & 36.0 & $46.9 \pm 0.8$ & 73.3 & 20.3 & 16.3 & \textbf{52.3} & \textbf{62.0} & 35.0 & 43.2 $\pm$ 1.3 \\
        $\pi_{0.5}$ & 56.8 & \textbf{26.2} & 38.0 & 94.1 & 36.4 & \textbf{36.8} & $48.1 \pm 0.7$ & 71.3 & 35.7 & 14.0 & 50.0 & 58.0 & 27.7 & 42.8 $\pm$ 1.5 \\
        VPP & 82.7 & 22.7 & 34.7 & 66.7 & 45.3 & 26.0 & $46.3 \pm 0.9$ & 44.0 & 60.0 & 22.7 &  30.7 & 52.0 & 45.3 & 44.7 $\pm$ 2.4 \\ 
        VPDD & 89.3 & 18.7 & 40.0 & 82.7 & 40.0 & 16.0 & $47.8 \pm 1.0$ & 44.7 & 53.3 & 20.7  & 32.0 & 36.0 & 44.0 & 38.5 $\pm$ 0.6 \\
        \hline 
        \rowcolor{gray!10} 
        \textbf{Ours} & \textbf{96.7} & 18.0 & \textbf{43.3} & \textbf{96.0} & \textbf{46.7} & 26.7 & \textbf{54.6} $\pm$ \textbf{1.1} & \textbf{76.0} & \textbf{85.3} & \textbf{36.7} & 36.0 & 50.0 & \textbf{40.7} & \textbf{54.1} $\pm$ \textbf{1.4} \\
        \hline 
      \end{NiceTabular}
    }
    \label{tab:sim_results}
\end{table*}
\section{Experiments}

% In this section, we conduct extensive experiments on both simulated and real-world robotic benchmarks to evaluate the effectiveness, efficiency, and robustness of our proposed framework, \textbf{Video2Act}. 
To evaluate the capability of Video2Act, we first compare the multi-task performance of our method with existing approaches on the RoboTwin 1.0 bimanual manipulation benchmark~\cite{Mu_2025_CVPR} and the RoboTwin 2.0 bimanual manipulation benchmark~\cite{chen2025robotwin} in Section~\ref{exp:simulation}. We then conduct ablation studies in Section~\ref{exp:ablation} to illustrate the effectiveness of the spatial and motion features extracted from the VDM and to analyze the relationship between system frequency ratios and manipulation performance. In Section~\ref{exp:real-world}, we present six real-world experiments to evaluate the robustness of our approach. In Section~\ref{exp:discussion}, we analyze how our method facilitates action learning through the visualization of action distributions and generalization scenarios.

\subsection{Simulation Experiment}
\label{exp:simulation}

\noindent\textbf{Simulation Benchmark.}
% To systematically evaluate our method, we conduct experiments in the \textbf{RoboTwin} simulation environment~\cite{Mu_2025_CVPR}, which is built upon the Sapien simulator. We adopt six various manipulation tasks from RoboTwin: \textit{block\_handover}, \textit{container\_place}, \textit{dual\_bottles\_pick\_easy}, \textit{dual\_bottles\_pick\_hard}, \textit{empty\_cup\_place}, and \textit{pick\_apple\_messy}. All tasks are performed using the ALOHA-AgileX dual-arm robot system. For data collection, expert demonstrations are automatically generated via motion planning based on key poses annotated by GPT-4. For each task, we generate 100 trajectories. Further details regarding the simulated manipulation experiments are provided in Appendix X.
To systematically evaluate our method, we conduct experiments in the RoboTwin simulation environment~\cite{Mu_2025_CVPR,chen2025robotwin}, which is built upon the Sapien simulator. Specifically, we adopt six manipulation tasks from \textbf{RoboTwin 1.0}: \textit{block handover}, \textit{container place}, \textit{dual bottles pick easy}, \textit{dual bottles pick hard}, \textit{empty cup place}, and \textit{pick apple messy}. We also train and evaluate on six tasks from \textbf{RoboTwin 2.0}: \textit{beat block hammer}, \textit{click bell}, \textit{hanging mug}, \textit{place object basket}, \textit{place shoe}, and \textit{turn switch}. All tasks are performed using the ALOHA-AgileX dual-arm robot system. For each task, we generate 100 expert demonstrations. Detailed setups are in Appendix A.1.
% Additional details of the simulation are shown in Appendix A.

% \paragraph{Baselines.}
% We compare \textbf{Video2Act} against five representing methods in robot imitation learning:
% \begin{itemize}
%     % \item \textbf{Diffusion Policy}~\cite{chi2024diffusionpolicy} formulates action learning as a conditional denoising process, iteratively refining action sequences from noise using a diffusion objective.
%     % \item \textbf{ACT}~\cite{Zhao-RSS-23} employs a Transformer-based CVAE to generate action chunks, reducing compounding errors and capturing variability in human demonstrations.
%     \item \textbf{Diffusion Policy}~\cite{chi2024diffusionpolicy} learns actions via conditional denoising, iteratively refining action sequences from noise.
%     \item \textbf{ACT}~\cite{Zhao-RSS-23} employs a transformer-based CVAE to generate coherent action chunks for bimanual manipulation.
%     \item \textbf{RDT-1B}~\cite{liu2024rdt} leverages a Robotics Diffusion Transformer conditioned on a pre-trained SigLIP encoder to model complex bimanual action distributions.
%     \item $\boldsymbol{\pi}_0$~\cite{Black2024pi0} augments a pre-trained VLM with an diffusion-based action expert module that uses conditional flow matching to generate high-frequency action sequences.
%     \item \textbf{Video Prediction Policy}~\cite{hu2024video} leverages one-step video prediction features extracted from a fine-tuned 
% UNet-based Stable Video Diffusion~\cite{blattmann2023stable} model to condition the diffusion-based action head.
% \end{itemize}
% \vspace{-2mm}
\noindent\textbf{Baselines.}
We compare \textbf{Video2Act} against seven representative methods in robot imitation learning:
Diffusion Policy~\cite{chi2024diffusionpolicy}; ACT~\cite{Zhao-RSS-23}; RDT-1B~\cite{liu2024rdt}; 
$\boldsymbol{\pi}_0$~\cite{Black2024pi0}; $\boldsymbol{\pi}_{0.5}$~\cite{Intelligence2025pi05}; VPDD~\cite{he2024learning} and VPP~\cite{hu2024video}. Notably, the first five baselines mainly rely on static image/VLM encoders (e.g., CLIP and SigLIP) without explicit long-horizon temporal video modeling, whereas VPDD and VPP explicitly exploit video diffusion features for action learning.

\noindent\textbf{Training and Evaluation Details.}
To ensure a fair comparison, all methods use a single model for multiple tasks and are trained and evaluated under the same configuration. For VPP and VPDD, we fine-tune Stable Video Diffusion on task demonstration videos and replace its action head with the pretrained diffusion head used in our method to ensure a fair comparison. During evaluation, we conduct 50 rollouts for each task with different seeds, repeating the evaluation three times per task and reporting the variance to ensure a robust comparison.

\noindent\textbf{Quantitative Results.} {As shown in Table~\ref{tab:sim_results}, Video2Act achieves an average success rate of 54.6\% and 54.1\% across the manipulation tasks in RoboTwin 1.0 and RoboTwin 2.0, respectively. This significantly outperforms previous state-of-the-art baselines $\pi_{0.5}$, $\pi_{0}$, and RDT by margins of 6.5\%, 7.7\%, and 9.7\% in RoboTwin 1.0, and by 11.3\%, 10.9\%, and 8.1\% in RoboTwin 2.0. Notably, Video2Act achieves superior performance on the majority of individual tasks, particularly excelling in scenarios requiring high-precision coordination and complex dynamics, such as \textit{block handover} and \textit{bottles hard} in RoboTwin 1.0, as well as \textit{hanging mug} and \textit{click bell} in RoboTwin 2.0.} By explicitly modeling spatio-motional representations, our approach gains a more accurate understanding of the manipulated object's structural and dynamic states, enabling more precise action generation. In contrast, methods relying on static visual features lack temporal perception, limiting their effectiveness in dynamic modeling during the manipulation process. While baselines like VPP and VPDD also leverage generative video models to capture temporal dynamics, they rely on implicitly learned raw representations that are not explicitly refined. Consequently, these representations tend to contain noisy and task-irrelevant information. {Video2Act achieves substantial improvements over VPP by margins of 8.3\% and 9.4\%, and similarly outperforms VPDD by margins of 6.8\% and 15.6\% across the diverse tasks in RoboTwin 1.0 and 2.0, respectively.} The results demonstrate that Video2Act can explicitly extract spatial structures and motion trajectories via Spatial Filtering and FFT, filtering out task-irrelevant information while preserving clean structural boundaries and coherent motion patterns.
% This results in more focused and robust representations, enhancing generalization and control precision in tasks requiring fine-grained reasoning.
% This results in more focused and robust representations, enhancing generalization and control precision in tasks requiring fine-grained reasoning.

% \begin{table*}[t!]
%     \centering
%     \caption{Real-World Generalization Results}
%     \label{tab:real_gen_results}
%     \renewcommand{\arraystretch}{1.2}
%     \newcommand{\stackcell}[1]{\begin{tabular}[c]{@{}c@{}}#1\end{tabular}}

%     \resizebox{\textwidth}{!}{%
%         \begin{tabular}{l|cccc|r|cccc|r}
%             \hline
%             \rowcolor{red!10!blue!10} 
%             \textbf{Methods} & \multicolumn{5}{c|}{\textbf{Pick Dual Flowers}} & \multicolumn{5}{c}{\textbf{Handover Cucumber}} \\
%             \hline
%              & \stackcell{Object\\Variation} & \stackcell{Background\\Variation} & \stackcell{Cluttered\\Table} & \stackcell{Light\\Variation} & \textbf{Mean} 
%              & \stackcell{Object\\Variation} & \stackcell{Background\\Variation} & \stackcell{Cluttered\\Table} & \stackcell{Light\\Variation} & \textbf{Mean} \\
%             \hline
%             RDT & & & & & & & & & & \\
%             $\pi_0$ & & & & & & & & & & \\
%             VPP & & & & & & & & & & \\
%             \hline
            
%             \rowcolor{gray!10}
%             \textbf{Video2Act} & XX.X & XX.X & XX.X & XX.X & \textbf{XX.X} & XX.X & XX.X & XX.X & XX.X & \textbf{XX.X} \\
%             \hline
%         \end{tabular}%
%     }
% \end{table*}
\begin{figure*}[t]
  \centering
  \includegraphics[width=0.98\textwidth]{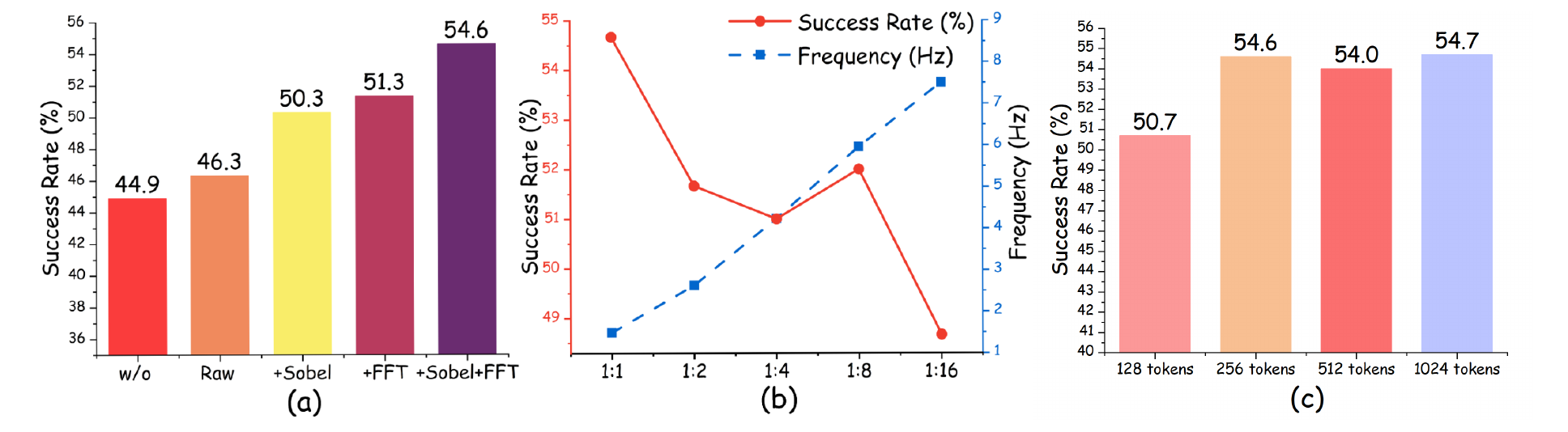}
  \caption{\textbf{Ablation Study.} We investigate (a) the effectiveness of spatio-motional feature extraction, (b) how the operating ratio influences success rate and action generation frequency and (c) the impact of different Q-Former token lengths on the success rate.}
  \label{fig:ablation}
  \vspace{-3mm}
\end{figure*}

\subsection{Ablation}
\label{exp:ablation}

To validate the effectiveness of each contribution, we conduct detailed ablation studies on the six simulated tasks in RoboTwin 1.0.

\noindent\textbf{Spatio-motional Feature Extraction in VDM.}
To evaluate the contribution of different visual representations extracted by the VDM, we compare five configurations:
(a) w/o VDM feature;
(b) raw VDM feature, using the unprocessed VDM features;
(c) +Sobel, using spatial-only features; (d) +FFT, using motion-only features; (e) +Sobel+FFT, combining both spatial and motion features. All variants use a fixed dual-system operating ratio of 1:1 to ensure a consistent comparison. As shown in Figure~\ref{fig:ablation}~(a), introducing the
Sobel-based spatial filter alone improves the task success rate by 4.0\%, while adding the FFT-based motion features alone yields a 5.0\% gain. 
When the two components are combined,
the model achieves its best performance (54.6\%), reaching an overall improvement of 8.3\%. These results indicate that our method effectively extracts spatial and temporal cues that provide complementary benefits: Sobel emphasizes structural information, while FFT captures motion consistency, and together they enhance the stability of action generation. In Appendix~B, we further investigate the impact of using latent features from different VDM transformer layers on manipulation performance.

\begin{figure*}[t]
    \centering
    \subfloat{
        \includegraphics[width=0.95\textwidth]{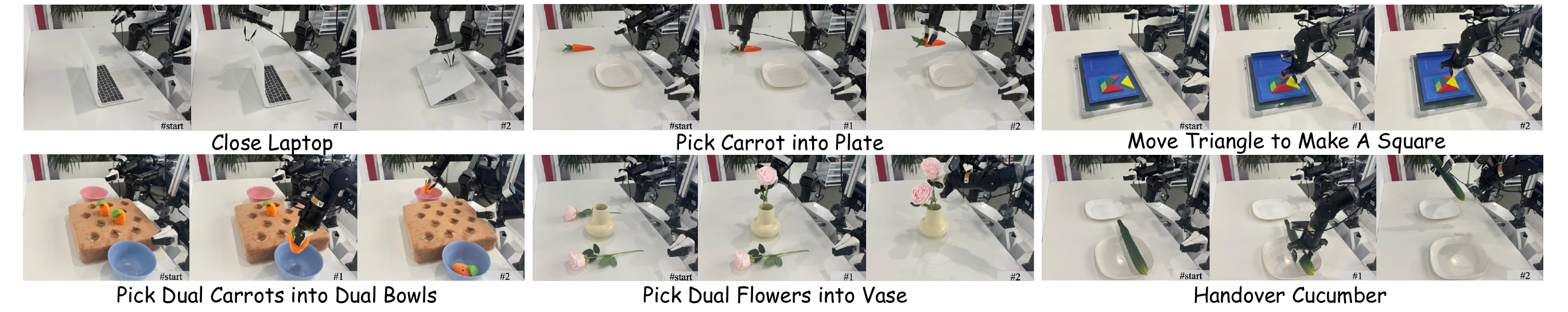}
        \label{fig5:Baseline}
    }
    \\
    
    \subfloat{
        \includegraphics[width=0.95\textwidth]{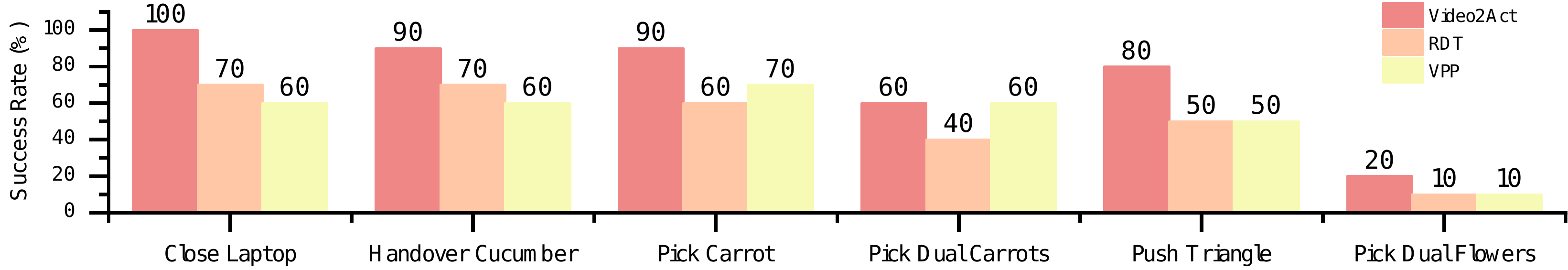}
        \label{fig:realworld_tasks}
    }
\vspace{-2mm}
    \caption{
     \textbf{Real-world experiment results across six manipulation tasks on the Agilex Cobot Magic platform.}
     All methods are trained on 100 demonstrations per task and compared against two closely related baselines, RDT and VPP. We report success rates over 10 rollouts per task under diverse tabletop configurations.
     }
    \label{fig5}
    \vspace{-5mm}
\end{figure*}

\noindent\textbf{Dual-System Operating Frequency Ratio.}
\label{exp:abaltion_ratio}
We further investigate how different operating frequency ratios affect performance. The ratio between the slow VDM component (System 2) and the fast action component (System 1) is varied from 1:1 to 1:16.
The ratio corresponds to different update intervals between the VDM latent features and the action module’s visual features. 
As shown in Figure~\ref{fig:ablation}~(b), the model achieves the best trade-off between accuracy and inference efficiency when the ratio is set to 1:8, resulting in a model inference frequency of 5.96 Hz. Applying Sobel and FFT operations in the latent space takes 0.56 ms and 1.33 ms, respectively, averaged over 50 runs, indicating that the additional overhead has a minor impact on overall speed.
Since we set the action chunk size to $H=64$, the overall system achieves real-time, high-frequency control (approximately 380 Hz).
Empirically, the 1:8 setting better exploits the inherent temporal modeling capability of the VDM. Ratios that either under-utilize (e.g., 1:2, 1:4) or over-utilize (e.g., 1:16) VDM temporal information result in suboptimal performance, whereas the 1:8 ratio achieves a more effective balance.
These results indicate that the VDM-extracted features inherently preserve temporal consistency across time steps, supporting asynchronous operation that improves inference efficiency without sacrificing accuracy.

\noindent\textbf{Spatial and Motion Former Token Length.}
As shown in Figure~\ref{fig:ablation}~(c), we conduct ablation studies with token lengths of 128, 256, 512, and 1024 for the Spatial and Motion Former across all simulation tasks, achieving success rates of 50.7\%, 54.6\%, 54.0\%, and 54.7\%, respectively. Performance degrades with 128 tokens, while longer token lengths achieve comparable results at the cost of increased computation. Therefore, to balance performance and computational efficiency while avoiding redundant information introduced by longer token sequences, we set the token length to 256.

\subsection{Real-World Experiment}
\label{exp:real-world}
\noindent\textbf{Dataset Collection.}
In our real-world robot experiment setup, we employ the Agilex Cobot Magic platform, which is equipped with a front-view and two wrist-view cameras. We perform six distinct manipulation tasks: 1) \textit{close laptop}, 2) \textit{pick carrot}, 3) \textit{pick dual carrots}, 4) \textit{pick dual flowers}, 5) \textit{handover cucumber} and 6) \textit{push triangle}, which involve various objects and action types. For each task, 100 demonstrations are collected via master-puppet teleoperation, with objects placed in varying positions on the table to ensure diversity. Hardware and data details are provided in Appendix A.2 and A.3.

\noindent\textbf{Training and Evaluation Details.}
All methods are trained to learn task-specific policies under the same configuration in simulation. We evaluate Video2Act against RDT~\cite{liu2024rdt}, $\pi_0$~\cite{Black2024pi0} and VPP~\cite{hu2024video}. The final checkpoint is used to perform 10 rollouts across varied tabletop positions. In our experiments, we use an ALOHA robot equipped with a NVIDIA 4090 GPU (24 GB VRAM). For power consumption, System 2 is run alongside the robot controller using half-precision (FP16) inference, requiring 21.04 GiB of GPU memory and 144.4 W of power. Since our robot platform does not support a multi-GPU setup, there is a cold-start latency of 587.9 ms before the first action generation (single VDM forward pass).
% We evaluate Video2Act against two closely related baselines, RDT~\cite{liu2024rdt}
% and VPP~\cite{hu2024video}. For evaluation, we perform 10 rollouts across varied tabletop positions.

\noindent\textbf{Quantitative and Qualitative Results.}
As shown in the second row of Figure~\ref{fig5}, Video2Act achieves superior real-world performance with an average success rate of \textbf{73.3\%} on the Agilex Robot, outperforming all baseline methods across six real-world tasks. Notably, Video2Act attains substantially higher success rates on bimanual manipulation tasks that require precise spatial reasoning and dynamic coordination. These results validate the effectiveness of our proposed Video2Act and its explicit spatio-motional representations in improving real-world robustness for complex manipulation tasks.
As shown in the first row of Figure~\ref{fig5}, Video2Act can accurately execute pick-and-place, articulated object manipulation, bimanual handover, and other precise manipulation tasks, highlighting the generality of our approach across diverse tasks. Additional visualizations, real-world evaluations, and failure case analyses are presented in Appendix C, D, and E, respectively.
% For example, in the Place Bottles
% at Rack task, our method attains a 70% success rate compared to π0 55%. Similarly, on the AlphaBot
% platform, FiS-VLA achieves a higher mean success rate of 74%, surpassing π0 61%. The greatest
% improvement is seen in the Fold towel and put task, which involves manipulating deformable objects.
% Qualitative results in Table 2 showcase FiS-VLA’s ability to execute diverse tasks across robots,
% including sequential bottle manipulation and blackboard erasing on Agilex, as well as fine-grained
% actions like pouring water on AlphaBot. These outcomes highlight the model’s effective coordination
% of high-level reasoning and low-latency control, enabling adaptive behavior in real-world settings.
% Additional visualizations and failure cases are provided in Appendix C and D, respectively.
% \vspace{-1mm}
\subsection{Analysis}
\label{exp:discussion}

\begin{figure*}[t]
  \centering
  \includegraphics[width=0.98\textwidth]{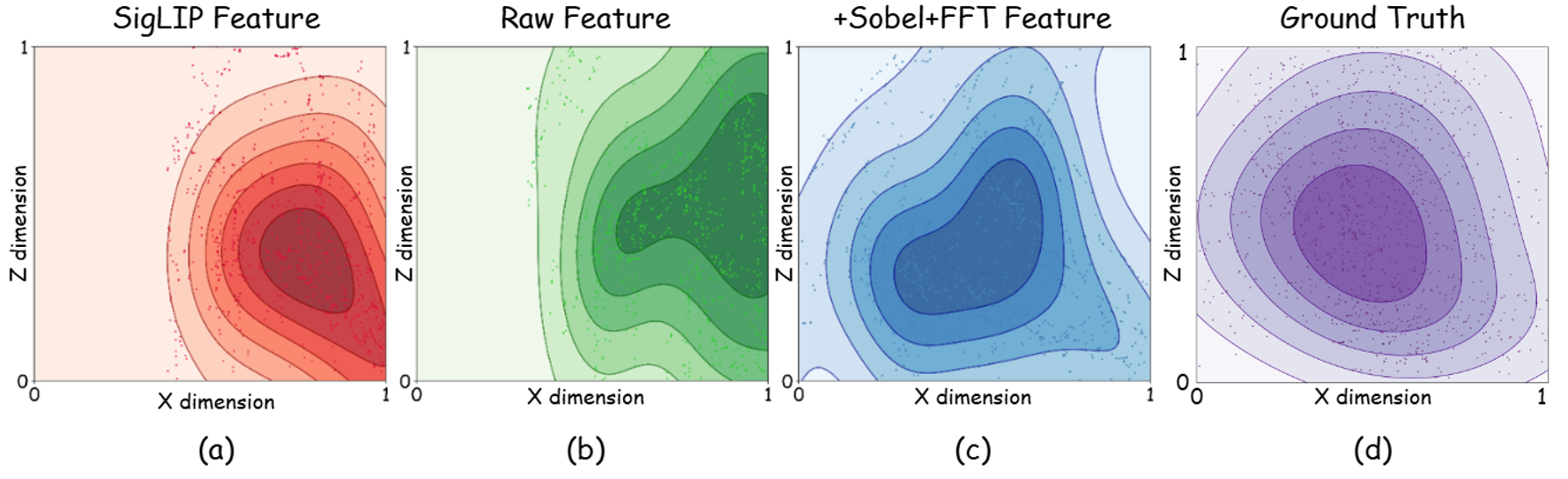}
  \vspace{-3mm}
  \caption{\textbf{Action Distribution Visualization.}
  We project the right-hand end-effector poses (X-Z dimensions) from 40 successful rollouts in the \textit{dual bottles pick hard} task. The comparison shows the learned distributions for: (a) SigLIP Feature; (b) Raw VDM Feature, and (c) +Sobel+FFT VDM Feature; (d) Ground Truth.}
  \vspace{-2mm}
  \label{fig:action_distribution}
\end{figure*}
\begin{figure*}[t]
  \centering
  \includegraphics[width=0.95\textwidth]{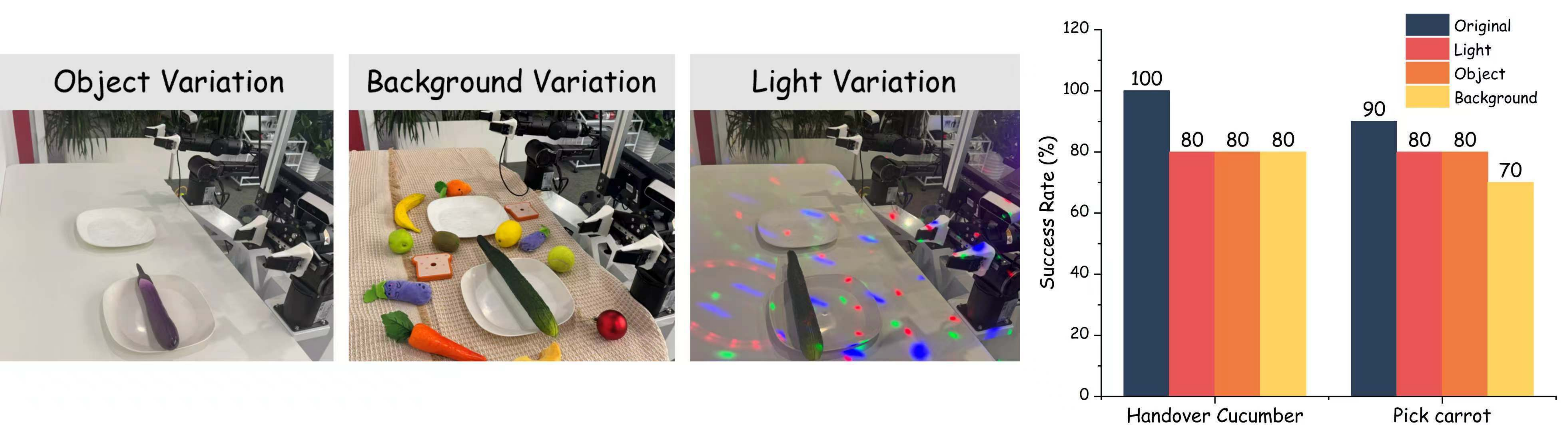}
    \vspace{-2mm}
    \caption{
    \textbf{Generalization results under unseen real-world scenarios.}
    We evaluate Video2Act on the \textit{pick dual flowers} and \textit{handover cucumber}
    tasks across three shifts, including object variation, background variation, and lighting variation.
    }
  \label{fig:generalization}
  \vspace{-3mm}
\end{figure*}

\noindent\textbf{Action Distribution Analysis.}
\label{exp:action_distribution}
To further analyze how the extracted spatio-motional features improve policy robustness and generalization, we visualize the action distribution learned by Video2Act. We conduct this study on the \textit{dual bottles pick hard} task, where the initial bottle positions are fixed to ensure controlled conditions. Following the task setup, the end-effector (EE) pose is projected along the world-frame y-axis, while the forward (x-axis) and upward (z-axis) directions remain unconstrained during grasping. For each configuration, we collect 40 successful trajectories and record the right-hand EE poses. We compare the action distributions along the X and Z dimensions for three model configurations:
(a) SigLIP feature; (b) raw VDM feature; (c) +Sobel+FFT feature, combining both spatial and motion features. We further report the distribution of expert demonstrations as ground truth in (d). As shown in Figure~\ref{fig:action_distribution}, the base policy (a), which relies only on static image features, exhibits a \textbf{narrowly concentrated} action distribution. Interestingly, directly adding the unprocessed VDM features (b) also demonstrates a relatively concentrated distribution, failing to fully learn the diverse trajectories. In contrast, conditioning on the refined +Sobel+FFT features (c) yields a markedly \textbf{broader} distribution that aligns most closely with the expert distribution in (d). The results demonstrate that our method captures object structural information and inter-object motion relationships, enabling it to learn a more diverse range of manipulation trajectories from demonstrations rather than overfitting to a narrow subset of the distribution.

% \begin{figure}[t]
%   \centering
%   \includegraphics[width=\columnwidth]{images/realworld_generalization_succ.png}
%   \caption{\textbf{123.}xxxx
%     }
%   \label{fig:generalization}
% \end{figure}

\noindent\textbf{Generalization Experiments.}
To further evaluate the zero-shot generalization capability of our method, we conduct experiments under several unseen configurations in the real-world tasks \textit{pick dual flowers} and \textit{handover cucumber}.
The unseen configurations include: 1) Object Variation, 2) Background Variation, and 3) Lighting Variation. As shown in \Cref{fig:generalization}, Video2Act maintains consistent success rates across all generalization scenarios.
The explicit spatial-aware representations enable our model to preserve robust object-structure understanding under varying object appearances and background contexts, while motion-aware representations provide stable dynamic priors that exhibit consistency to visual disturbances. The performance demonstrates that our approach has the potential to capture essential manipulation concepts that generalize to novel environments.

% As shown in \Cref{fig:generalization}, Video2Act maintains consistent success rates in all generalization scenarios. The explicit spatial-aware representations enable our model to sustain robust object localization and boundary detection under varying object appearances and background contexts, while the motion-aware representations provide stable dynamic priors that demonstrate strong resilience to visual distractions and lighting variations. This stability aligns with our core insight that by explicitly encoding structural and motion consistency, VDM can learn transferable spatio-temporal patterns that remain effective beyond the training distribution. The consistent performance across diverse real-world variations validates that our approach effectively captures essential manipulation concepts that generalize well to novel environments.
\section{Conclusion}
\label{sec:conclusion}

In this paper, we presented Video2Act, an asynchronous dual-system VLA framework that leverages the spatio-temporal representations of video diffusion models to enhance robotic policy learning. Our systematic analysis revealed that VDM features naturally encode spatial- and motion-aware representations that remain robust to robot motion and viewpoint changes. Building on this insight, we proposed explicit spatial and motion representation extraction methods using Spatial Filtering Operators and FFT, which effectively capture what to manipulate and how to move. 
Meanwhile, an asynchronous dual-system strategy is introduced, in which the VDM is designated as the slow perceptual System 2, and the action head is assigned as the fast execution System 1, thereby enabling high-frequency and stable action generation. Extensive experiments in both simulation and real-world environments demonstrate that Video2Act achieves SOTA performance across various manipulation tasks. 
% For the limitation, Video2Act inherits the computational overhead of VDMs, and leveraging deeper or more numerous VDM latent features increases inference cost. While our asynchronous dual-system design partially alleviates this issue, developing more efficient ways to utilize VDM representations remains an important direction for future work.

{
    \small
    \bibliographystyle{ieeenat_fullname}
    \bibliography{main}
}

\clearpage
\clearpage
\setcounter{page}{1}
\maketitlesupplementary
\appendix

\definecolor{condA}{RGB}{255,255,255}
\definecolor{condB}{RGB}{243,243,243}
\definecolor{condC}{RGB}{229,229,229}
\definecolor{avggray}{RGB}{246,246,246}

We provide additional details, as well as quantitative and qualitative results of our Video2Act in this supplementary material. The outline is shown below.

\begin{itemize}
    \item \textbf{A. Additional Experimental Setup and Data Details (Appendix~\ref{apsec:A})}
    \begin{itemize}
        \item Simulation Setup and Data Collection
        \item Real-World Robot Hardware Setup
        \item Real-World Data Collection
    \end{itemize}

    \item \textbf{B. Additional Ablation Study (Appendix~\ref{apsec:B})}
    \begin{itemize}
        \item Spatial Filter Choice Ablation
        \item Layer-Depth Configuration Ablation
        \item Dual-Stream Input Window Size ($T_s$, $T_l$) and Resolution Ablation
        \item Further Action Distribution Analysis
    \end{itemize}

    \item \textbf{C. Additional Visualizations (Appendix~\ref{apsec:C})}
    \begin{itemize}
        \item Real-World Grad-CAM Visualizations
        \item Simulation Qualitative Results
        \item Real-World Qualitative Results
    \end{itemize}

    \item \textbf{D. Additional Real-World Evaluations (Appendix~\ref{apsec:D})}
    \begin{itemize}
        \item Edge-Case Lighting and Zero-Shot Object Evaluation
    \end{itemize}

    \item \textbf{E. Failure Analysis (Appendix~\ref{apsec:E})}
    \begin{itemize}
        \item Failure Case Analysis and Visualization
    \end{itemize}

    \item \textbf{F. Additional Quantitative Generalization Experiments (Appendix~\ref{apsec:F})}
    \begin{itemize}
    \item Generalization Performance with Light and Background Randomization in RoboTwin 2.0
    \end{itemize}
\end{itemize}

\section{Additional Experimental Setup and Data Details}
\label{apsec:A}

In this section, we provide additional details on the simulation configuration, real-world robot hardware, and data collection procedures that complement the descriptions in the main text.

\begin{figure}[t]
  \centering
  \includegraphics[width=0.6\columnwidth]
  {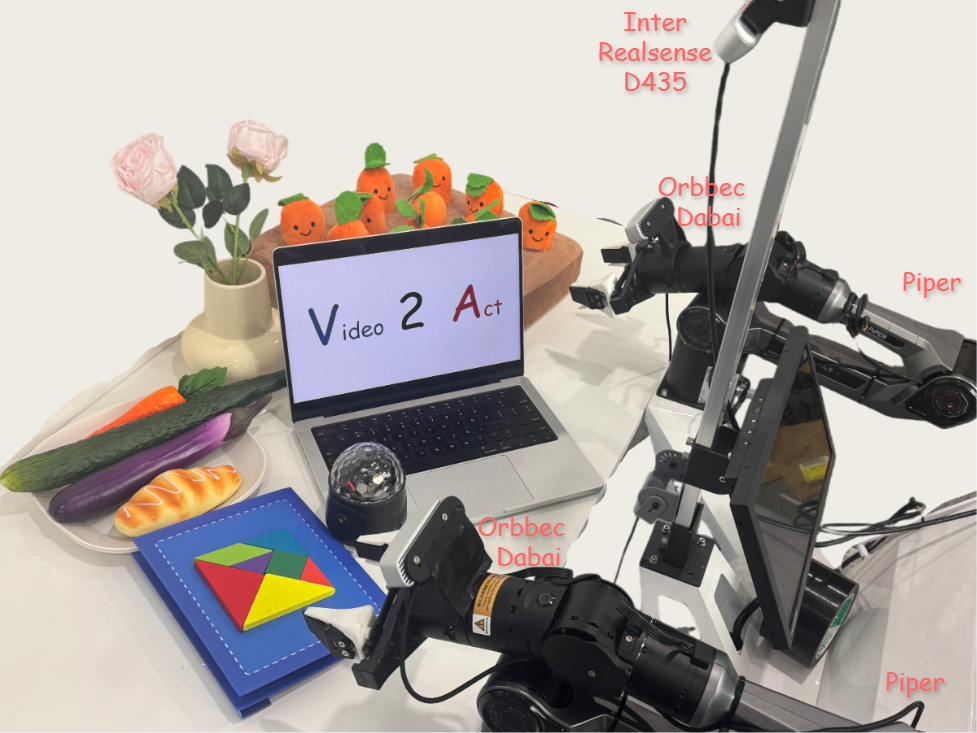}
  \caption{\textbf{Real-world robot setup and experimental assets.} We utilize the Agilex Cobot Magic platform with four 6-DoF Agilex Piper robotic arms, equipped with an Intel RealSense D435 head camera and two Orbbec Dabai wrist cameras, along with the object assets used across all real-world tasks.}
  \label{fig:asset}
\end{figure}

\subsection{Simulation Setup and Data Collection}
\label{apsec:A1}
In RoboTwin~\cite{Mu_2025_CVPR}, demonstrations are automatically generated via a generative digital twin pipeline. First, diverse 3D assets are created from single 2D images using generative foundation models and annotated with functional axes and contact points. Subsequently, GPT-4~\cite{achiam2023gpt4} decomposes the manipulation tasks and infers spatial constraints based on these annotations. Finally, these constraints are translated into executable code that drives a motion planner to solve for collision-free, kinematically feasible trajectories. For each task, we generate 100 expert demonstrations using this automated framework. Detailed descriptions of twelve simulation tasks are as follows:
\begin{description}
    \item[\textit{1. block handover.}] A long block is initialized on the left side of the table. The left arm grasps the upper side of the block and performs a handoff to the right arm, which then places the block onto a blue mat on the right side. This task requires high-level inter-arm coordination to ensure a stable transition of the object between grippers without dropping it.

    \item[\textit{2. container place.}] Random containers, such as cups or bowls, are placed arbitrarily on the table. The robot must identify the container's location to select the appropriate arm (left or right) and move the container into a fixed plate. This task tests the model's ability to handle object diversity and make dynamic arm-selection decisions based on spatial distribution.

    \item[\textit{3. dual bottles pick easy.}] A red bottle and a green bottle are placed upright on the left and right sides of the table, respectively. The robot utilizes both arms simultaneously to grasp and lift the two bottles to a designated location. This task demands synchronized control of both end-effectors to execute parallel manipulation actions efficiently.

    \item[\textit{4. dual bottles pick hard.}] Similar to the easy setting, but the bottles are initialized with random postures (e.g., lying down) rather than standing upright. The model must perceive the complex 6D poses of the objects and perform precise orientation adjustments to align the grippers for a successful dual-arm grasp.

    \item[\textit{5. empty cup place.}] An empty cup and a coaster are randomly placed on either the left or right side of the table. The robot must grasp the cup and accurately place it onto the coaster. This task requires fine-grained spatial reasoning to align the cup with the coaster's surface while avoiding collisions.

    \item[\textit{6. pick apple messy.}] An apple is placed on the table surrounded by four random distractor items. The robot is required to identify, grasp, and lift the apple amidst the clutter. This task challenges the model's visual robustness and ability to plan collision-free trajectories in a cluttered environment.

    \item[\textit{7. beat block hammer.}] A hammer and a block are initialized on the table. The robotic arm must grasp the hammer and use it as a tool to strike the block. This task requires the model to understand tool use and dynamic physical interaction, demanding precise trajectory and velocity control to successfully impact the target object.

    \item[\textit{8. click bell.}] A desk bell is placed on the table. The robot must approach the bell and precisely press its top center button to ring it. This task tests the model's fine-grained localization and precision manipulation capabilities, requiring accurate interaction with a specific, small functional part of an object.

    \item[\textit{9. hanging mug.}] A mug and a mug rack are initialized on the table. The left arm grasps the mug, rotates it to a suitable orientation, and places it in the middle of the table. Subsequently, the right arm grasps the reoriented mug and carefully hangs it onto the rack. This complex task demands long-horizon, multi-step planning and asymmetric bimanual coordination, involving sequential state preparations and precise 6D pose alignment for the final hanging action.

    \item[\textit{10. place object basket.}] A target object and a basket are randomly placed in the workspace. One arm is designated to grasp the object and place it into the basket. After successful placement, the opposite arm grasps the basket and moves it to a new location. This task evaluates sequential multi-arm collaboration and spatial reasoning, requiring the robot to smoothly transition between manipulating a single object and transporting a container.

    \item[\textit{11. place shoe.}] A shoe and a target mat are initialized on the table. The robotic arm must grasp the irregularly shaped shoe and accurately place it onto the mat. This task assesses the model's ability to perceive and manipulate everyday objects with complex geometries and perform precise placement requiring specific spatial alignment.

    \item[\textit{12. turn switch.}] A switch mechanism is located in the workspace. The robotic arm is required to navigate to the switch and toggle its state by pressing it. This task challenges the model's fine motor control and its ability to identify and interact with functional mechanisms, demanding precise end-effector positioning.
\end{description}

\subsection{Real-World Robot Hardware Setup}
\label{apsec:A2}

In our real-world robot experiments, we employ the Agilex Cobot Magic platform equipped with four 6-DoF Agilex Piper robotic arms. Both the inference and execution puppet arms are equipped with parallel grippers featuring an 85 mm stroke. The robotic arms operate under joint position control, with the motion range of each joint detailed in Table~\ref{tab:joint_specs}. The robot is integrated with three cameras: a head-view Intel RealSense D435 camera~\cite{intel2026realsense} mounted on the head, and two wrist-view Orbbec Dabai cameras—one attached to the left wrist and the other to the right wrist. The specific parameters of the two cameras are shown in Table~\ref{tab:camera_configs}. The head camera provides a global perspective for environmental perception, while the two wrist cameras offer localized visual feedback for fine-grained manipulation tasks performed by the respective arms. The overall hardware configuration and real-world experimental assets are shown in Figure~\ref{fig:asset}. All algorithms utilize RGB information from all three cameras.

\begin{table}[t]
    \centering
    
    \caption{Agilex Piper Arm Joint Specifications}
    \label{tab:joint_specs} 
    \begin{tabular}{ccc} 
        \toprule
        Joint Name & Range & Maximum Speed \\
        \midrule
        J1 & $-154^\circ \sim 154^\circ$ & 180$^\circ$/s \\
        J2 & $0^\circ \sim 195^\circ$ & 195$^\circ$/s \\
        J3 & $-175^\circ \sim 0^\circ$ & 180$^\circ$/s \\
        J4 & $-100^\circ \sim 112^\circ$ & 225$^\circ$/s \\
        J5 & $-75^\circ \sim 75^\circ$ & 225$^\circ$/s \\
        J6 & $-170^\circ \sim 170^\circ$ & 225$^\circ$/s \\
        \bottomrule
    \end{tabular}
    
    \vspace{0.5cm} %
    
    \caption{Configurations for cameras}
    \label{tab:camera_configs} 
    \begin{tabular}{lcc}
        \toprule
        Parameter & Head Camera & Wrist Camera \\
        \midrule
        Resolution (H$\times$W) & 640 $\times$ 480 & 640 $\times$ 480 \\
        FOV (H$\times$W) & 56$^\circ$ $\times$ 43$^\circ$ & 67$^\circ$ $\times$ 52$^\circ$ \\
        Frequency & 30 fps & 30 fps \\
        \bottomrule
    \end{tabular}
\end{table}

\subsection{Details of Real-World Data Collection}
\label{apsec:A3}

% \noindent\textbf{Self-collected and Real-world Dataset.} 
Building upon our robot hardware setup, we collect six challenging real-world tasks, comprising three single-arm tasks and three bimanual tasks. For each task, 100 demonstrations were collected via master–puppet teleoperation~\cite{fu2024mobile}. Each demonstration comprises time-synchronized recordings of the puppet arm's joint positions and RGB video streams from three fixed perspectives. To ensure data diversity, objects were placed in varying positions on the table. Detailed descriptions of six robotic tasks are as follows:
\begin{description}
    \item[\textit{1. close laptop.}] The robot uses its arm to close the laptop’s opened folding screen. This task requires precise spatial perception to locate the screen and the hinge, as well as controlled force exertion to avoid damaging the screen during contact. The motion trajectory must be smooth and consistent to ensure stable manipulation of the articulated object.

    \item[\textit{2. pick carrot.}] The robot moves its arm to the carrot’s position, grasps the carrot, transfers it above the plate, and releases it into the plate. The task relies heavily on visual localization of the carrot and the plate, while the grasping and releasing actions demand accurate pose prediction and gripper control. The motion must be stable to prevent the carrot from rolling or falling during transfer.

    \item[\textit{3. pick dual carrots.}] The robot places the right carrot into the right bowl, then the left carrot into the left bowl (with carrot position variations). This bimanual task requires coordinated motion planning and spatial reasoning to handle positional variations. The model must perceive the geometric relationship between each carrot and its target bowl, and execute sequential actions without interference between the two arms.

    \item[\textit{4. pick dual flowers.}] The robot first grasps the right flower and inserts it into the vase, then grasps the left flower and inserts it into the vase. The insertion process demands fine-grained spatial awareness to align the flower stem with the vase opening. The model must also avoid colliding with the first flower when inserting the second, highlighting the need for dynamic trajectory adjustment.

    \item[\textit{5. handover cucumber.}] The robot grasps the cucumber from the plate with its left hand, transfers it to the right hand, and places it into the right plate. This task involves inter-arm coordination and precise timing during the handover phase. The model must ensure a stable grip transition and avoid dropping the cucumber, relying on both spatial and motion-aware representations to synchronize the dual-arm actions.

    \item[\textit{6. push triangle.}] The robot pushes a triangle to complete a tangram pattern into a square. The task requires understanding of geometric relationships and spatial composition. The pushing motion must be carefully planned to align the triangle with the existing pattern, involving both structural perception and trajectory optimization to achieve the target configuration.
\end{description}

\section{Additional Ablation Study}
\label{apsec:B}

\subsection{Spatial Filter Choice Ablation}
\label{apsec:B1}

Spatial filtering in Video2Act aims to extract stable structure-aware cues from the VDM latent features. We compare Sobel~\cite{sobel1968} with two representative alternatives: a second-order Laplacian~\cite{gonzalez2008digital} operator and a high-quality first-order Scharr~\cite{scharr2000optimal} operator. We keep the motion branch (FFT~\cite{cooley1965algorithm}) fixed and only vary the spatial filter.

\begin{figure}[t]
  \centering
   \hspace*{-0.5cm}
  \includegraphics[width=\columnwidth]{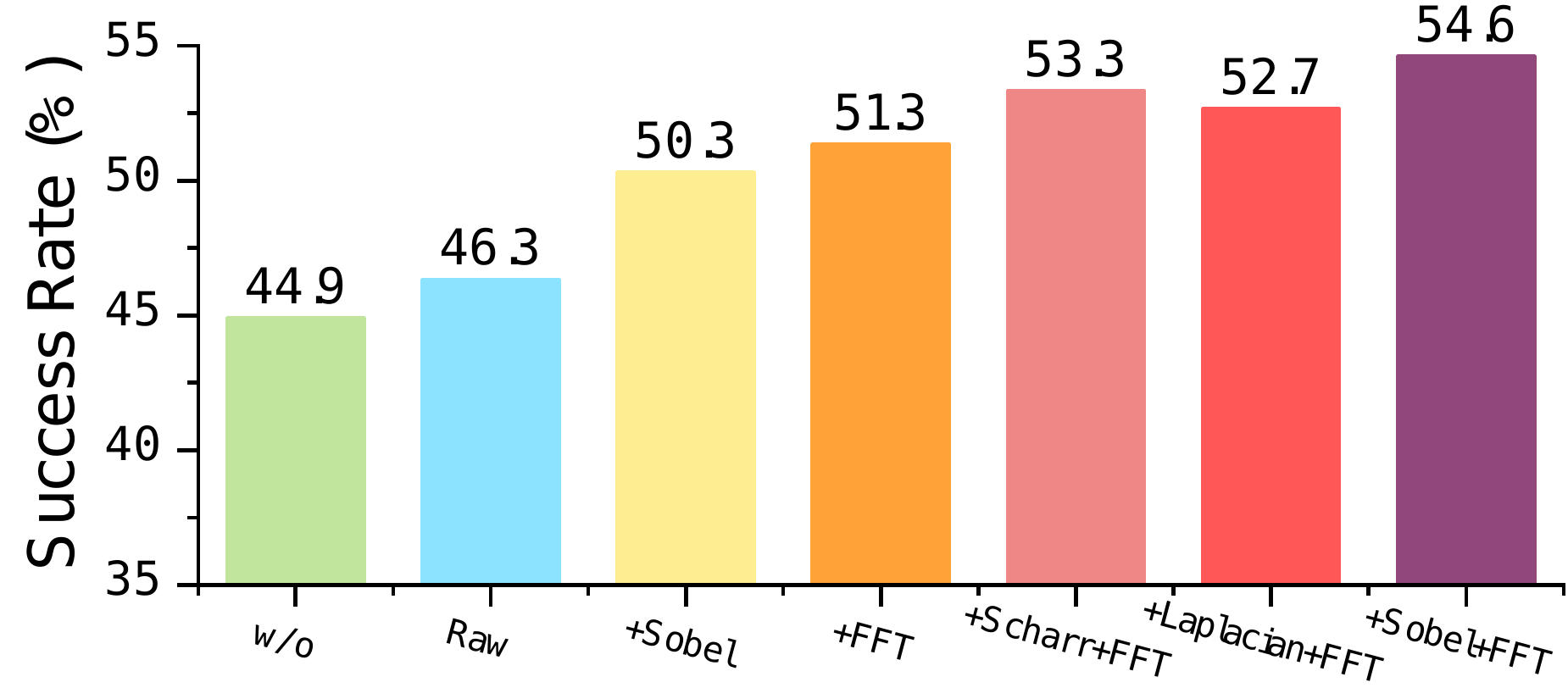}
  \caption{\textbf {Spatial filter ablation.} We add +Scharr+FFT and +Laplacian+FFT to our spatial filtering module and evaluate their six-task average success rates on RoboTwin under the same settings as in the main text.}
  \label{appendix:filter}
\end{figure}

As shown in Figure~\ref{appendix:filter}, we can see that +Scharr+FFT reaches an average success rate of 53.3 \%, while Laplacian reaches 52.7 \%, indicating that spatial filtering indeed benefits manipulation and that Scharr performs very similarly to Sobel. The slightly lower score of Laplacian may be due to its isotropic second-order formulation, which tends to amplify high-frequency noise and weaken directional edge responses. In contrast, first-order operators such as Sobel and Scharr emphasize gradient orientation and object contours more effectively, producing spatial cues that are more aligned with manipulation-relevant boundaries. This suggests that preserving directional edge information is more beneficial than enforcing rotational isotropy when extracting structure from VDM latents. 

\subsection{Layer-Depth Configuration Ablation}
\label{apsec:B2}

The video diffusion model used in Video2Act follows the dual-stream/single-stream hierarchy of the Hunyuan VDM architecture~\cite{kong2024hunyuanvideo}. 
The first 20 \textit{dual-stream transformer blocks} perform multimodal self-attention among the SigLIP image embeddings, text embeddings, and latent features, enabling early fusion of visual and instruction information. 
The subsequent 40 \textit{single-stream transformer blocks} apply cross-attention between the latent features and the fused multimodal tokens, progressively enriching the representation with deeper semantic structure.

While this hierarchical design is well suited for video generation, it is not clear how much depth is actually needed for extracting the spatio-motional cues required by our policy, or whether using more cross-attention blocks offers tangible benefit beyond increasing inference latency. 
To investigate this behavior, we perform a structured ablation aligned with the natural stage boundaries of the dual-stream and single-stream blocks.

We vary the depth across these two stages as follows:
\begin{itemize}[leftmargin=1.2em]
    \item 5 dual-stream transformer blocks
    \item 20 dual-stream transformer blocks
    \item 20 dual-stream blocks + 5 single-stream blocks (ours)
    \item 20 dual-stream blocks + 20 single-stream blocks
    \item 20 dual-stream blocks + 40 single-stream blocks
\end{itemize}

We evaluate how the average success rate and the System-2 perceptual latency—measured as the combined time of the VDM forward pass and feature-processing pipeline—change across these configurations. The results show that adding a small number of single-stream blocks on top of the 20 dual-stream blocks provides the best trade-off between depth and performance. The 20 DS + 5 SS configuration (ours) achieves the highest success rate of 54.6\%, indicating that a lightweight amount of cross-attention is sufficient for extracting the spatio-motional cues needed for policy learning. Increasing the number of single-stream blocks to 20 or 40 does not consistently improve performance and instead leads to diminishing or fluctuating gains, suggesting that deeper cross-attention introduces additional latency without contributing meaningful new information. 

\begin{table}[t]
\centering
\begin{minipage}{\columnwidth}
\centering
\caption{
Layer-depth configuration ablation using different numbers of 
dual-stream (DS) and single-stream (SS) transformer blocks.
}
\label{tab:layer_depth_ablation}
\begin{tabular}{lcc}
\toprule
Configuration & Success (\%) & Latency (ms) \\
\midrule
5 DS                  & 51.3 & 253.6 \\
20 DS                 & 52.7 & 488.4 \\
20 DS + 5 SS (ours)   & \textbf{54.6} & 587.9 \\
20 DS + 20 SS         & 52.3 & 886.5 \\
20 DS + 40 SS         & 54.0 & 1284.6 \\
\bottomrule
\end{tabular}
\end{minipage}
\end{table}

\begin{table}[t]
    \centering
    \caption{\textbf{Ablation on input window sizes and resolution.} We report the six-task average success rate on RoboTwin 1.0. The first row represents our default configuration.}
    \label{tab:ts_tl_res_ablation}
    \footnotesize
    \begin{tabular}{lcccc}
        \toprule
        Variant & $T_s$ & $T_l$ & Resolution & Success (\%) \\
        \midrule
        \rowcolor{gray!10}
        Default (Ours) & 2 & 16 & $512\times768$ & \textbf{54.6} \\ 
        \midrule
        Increased short window & 8 & 16 & $512\times768$ & 55.3 \\
        Reduced long window & 2 & 8 & $512\times768$ & 48.7 \\
        Lower resolution & 2 & 16 & $256\times256$ & 52.3 \\
        \bottomrule
    \end{tabular}
\end{table}

\subsection{Dual-Stream Input Window Size ($T_s$, $T_l$) and Resolution Ablation}
\label{apsec:B3}
% high-resolution input $I^{H}_{t-T_s:t} window size T_s, long-horizon input window size T_l $I^{L}_{t-T_l:t}

For the choice of window size $T_s$ and $T_l$, we further ablate $T_s$ and $T_l$ together with input resolution. In the main paper's results, we choose $T_s = 2$ and $T_l = 16$. We additionally evaluate $T_s = 2$ and $T_s = 8$, achieving comparable performance (54.6\% and 55.3\%, respectively). We further study the effect of high-resolution input $I^H$ resolution by reducing it from $[512, 768]$ to $[256, 256]$, leading to average performance drop to 52.3\%, indicating that higher resolution is more important for spatial filtering. For the long-horizon window size, $T_l = 8$ performs significantly worse than $T_l = 16$ (48.7\%). Overall, we set $T_s = 2$ and $T_l = 16$ to provide a trade-off between inference efficiency and accuracy.

\begin{figure}[t]
  \centering
  \includegraphics[width=\columnwidth]
  {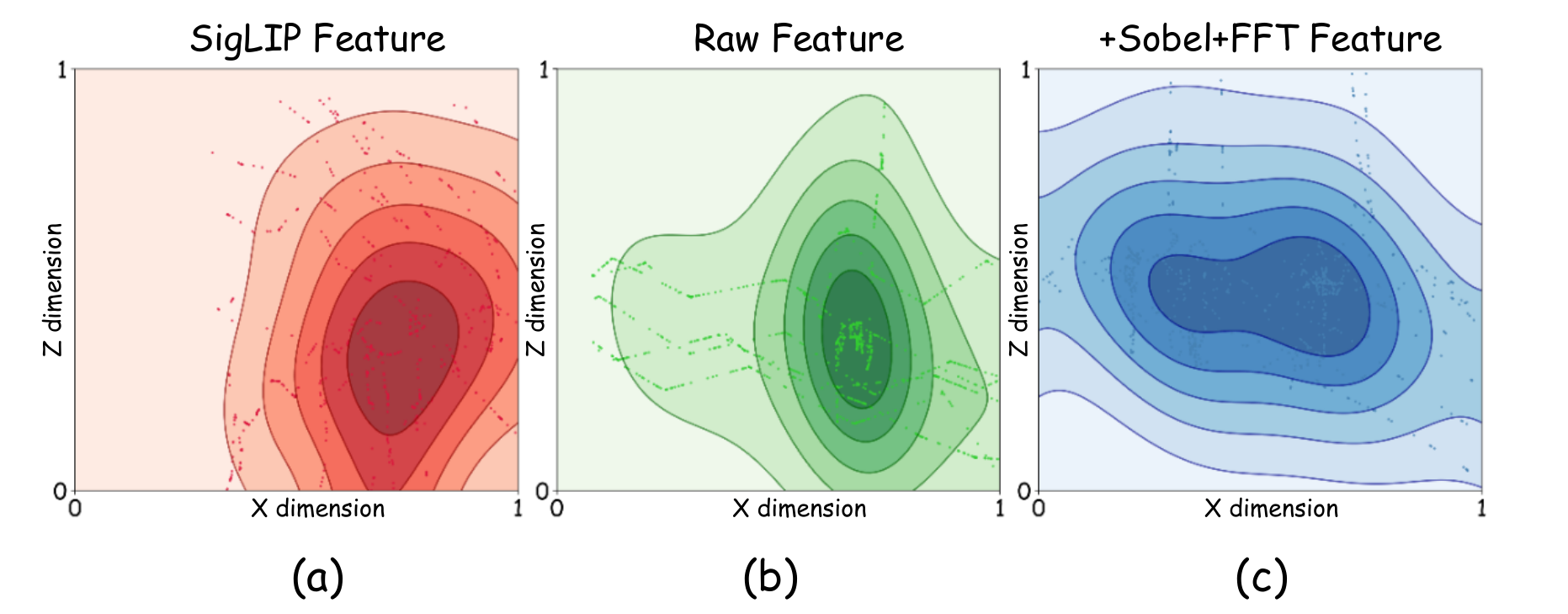}
  \caption{\textbf{Action Distribution Visualization.} 
  We project the left-hand end-effector poses (X-Z dimensions) from 40 successful rollouts in the \textit{dual bottles pick hard} task. The comparison shows the learned distributions for: (a) SigLIP Feature; (b) Raw VDM Feature, and (c) +Sobel+FFT VDM Feature.
  }
  \label{fig:action_distribution2}
\end{figure}

\begin{figure*}[t]
  \centering
  \includegraphics[width=\textwidth]{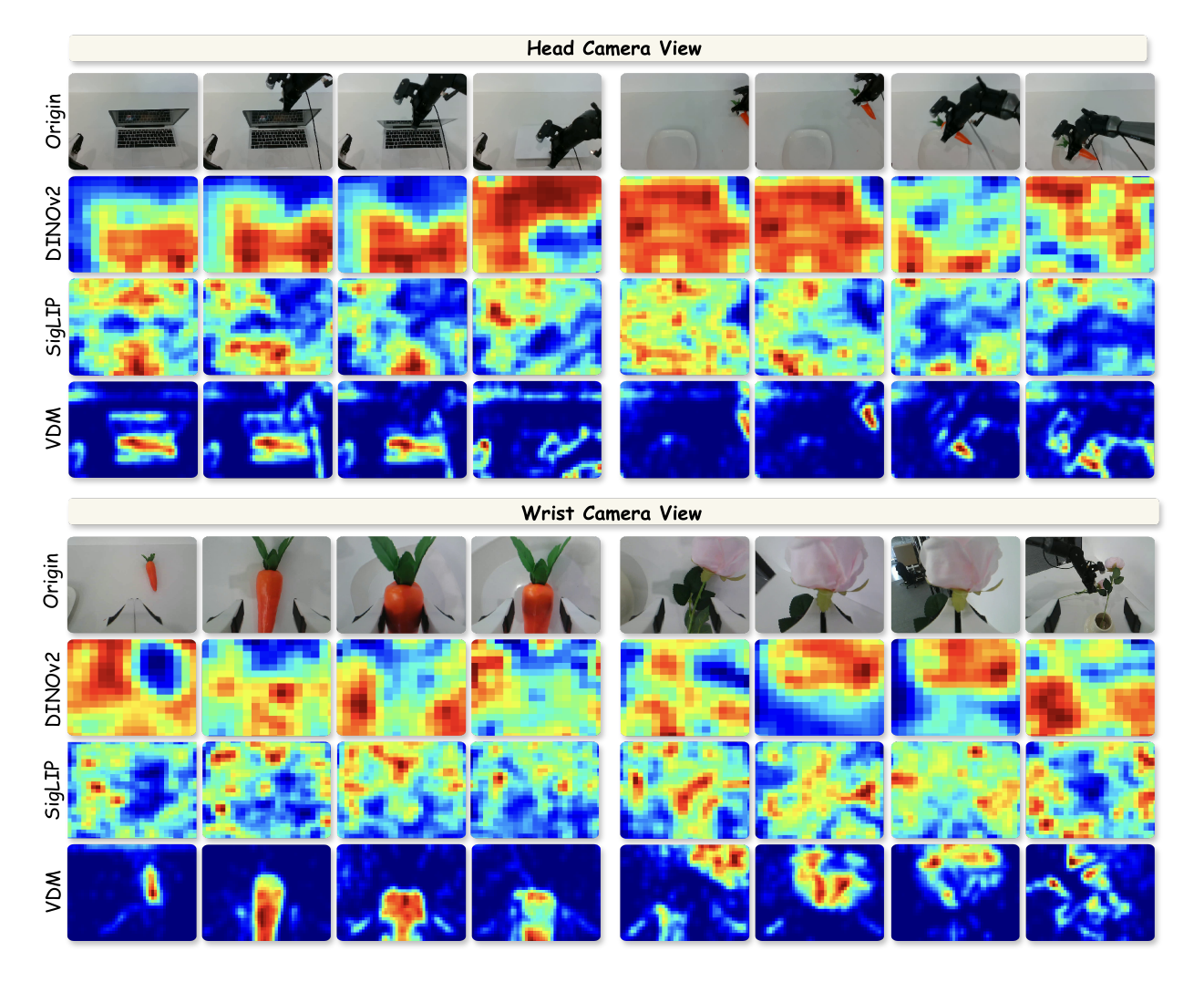}
  \caption{\textbf{Grad-CAM comparison on real-world scenarios.} We compare DINOv2, SigLIP, and our VDM-based representation on \textit{close laptop} and \textit{pick and place carrot} from the head camera view, and on \textit{pick and place carrot} and \textit{pick dual flowers} from the wrist camera view. DINOv2 and SigLIP exhibit scattered and unstable attention that often drifts across frames, whereas the VDM representation consistently maintains an object-centric focus.}
  \label{fig:real_gradcam}
\end{figure*}

\begin{figure*}[t]
  \centering
  \includegraphics[width=\textwidth]{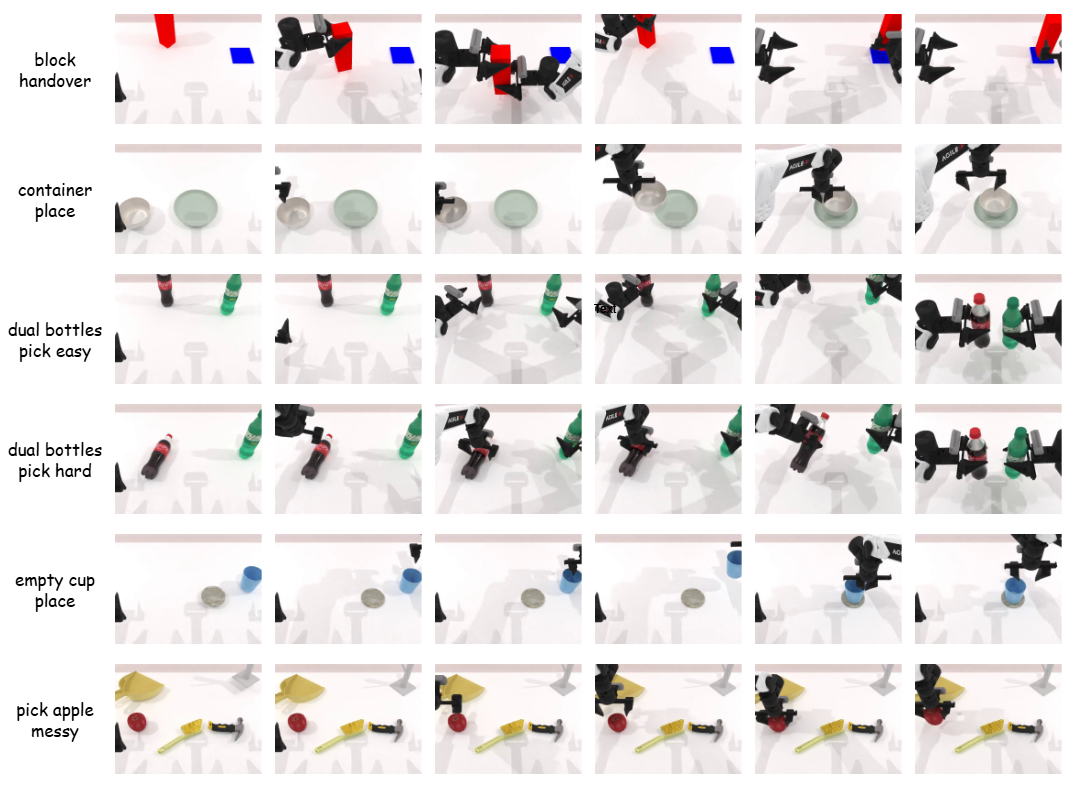}
  \caption{\textbf{Robot execution progress in RoboTwin1.0 simulation tasks.}We visualize key frames of the robot’s execution process from a static exterior view in simulation tasks.}
  \label{fig:simulation1}
\end{figure*}
\begin{figure*}[t]
  \centering
  \includegraphics[width=\textwidth]{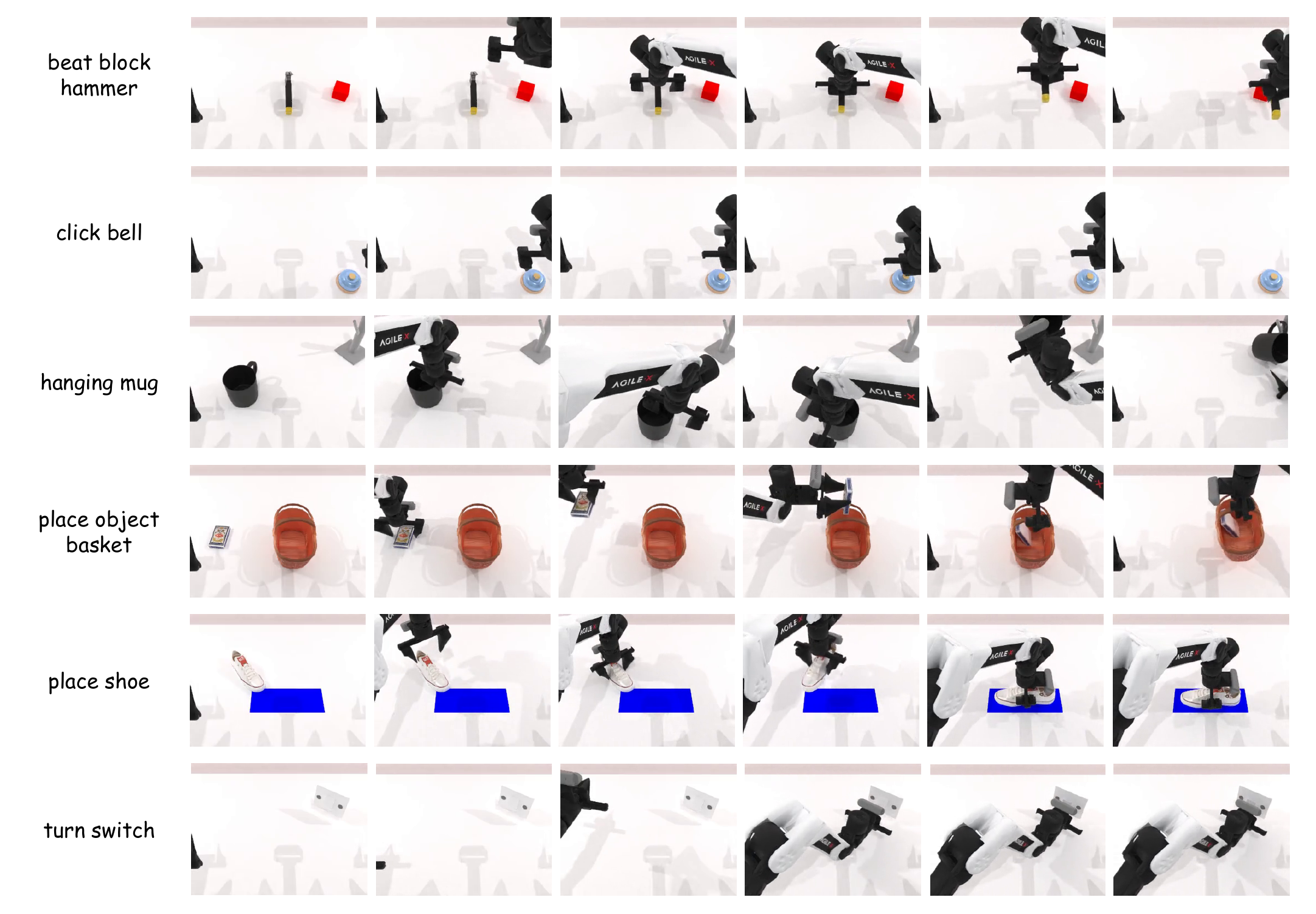}
  \caption{\textbf{Robot execution progress in RoboTwin2.0 simulation tasks.}We visualize key frames of the robot’s execution process from a static exterior view in simulation tasks.}
  \label{fig:simulation2}
\end{figure*}

\begin{figure*}[t]
  \centering
  \includegraphics[width=\linewidth]
  {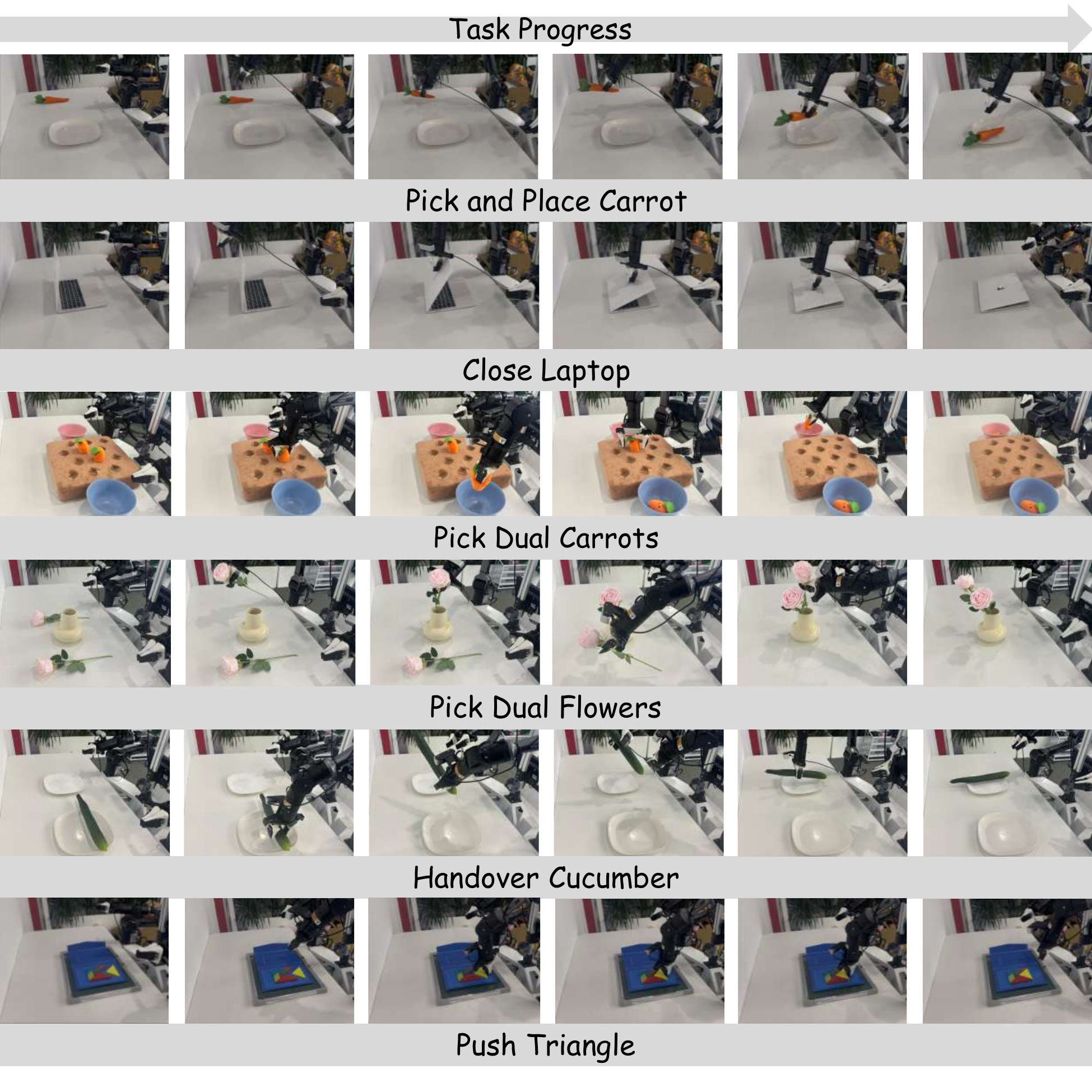}
  \caption{\textbf{Robot execution progress in real-world tasks.}We visualize key frames of the robot’s execution process from a static exterior view in real-world tasks.}
  \label{fig:detailed_tasks}
\end{figure*}

\begin{figure*}[t]
    \centering
    \includegraphics[width=\textwidth]{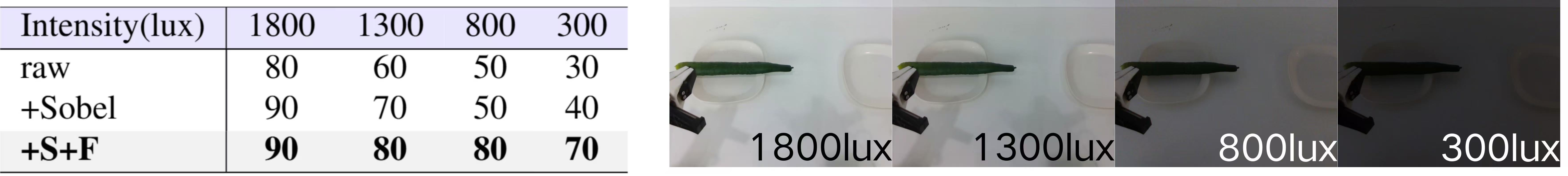}
    \caption{\textbf{Edge-case evaluation under varying illumination levels on the \textit{handover cucumber} task.}
    We report the performance of Raw, +Sobel, and +Sobel+FFT across four illumination levels.}
    \label{fig:lightning_edge}
    \vspace{-1em}
\end{figure*}

\begin{figure*}[t]
  \centering
  \includegraphics[width=\linewidth]
  {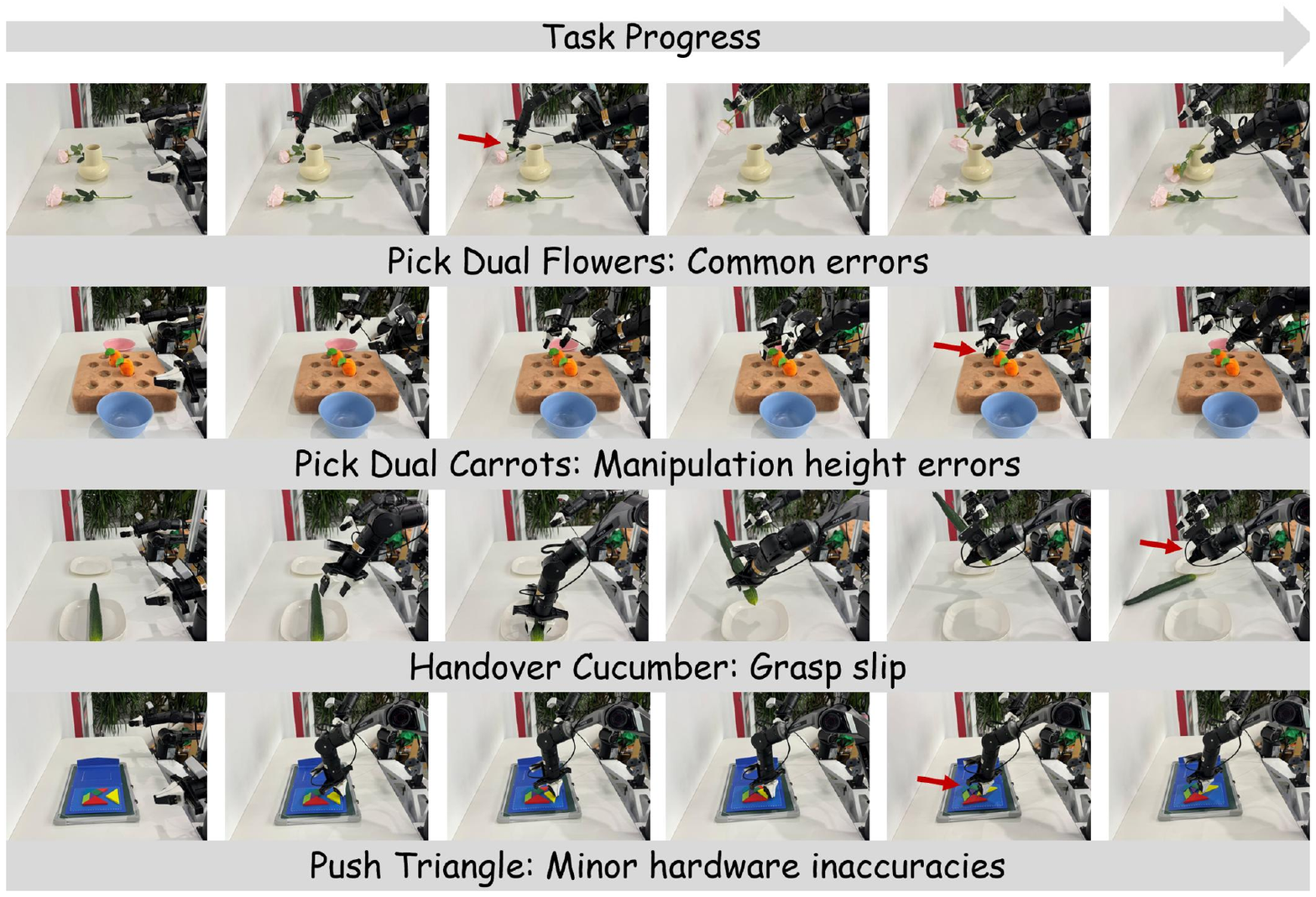}
  \caption{\textbf {Failure cases in real-world tasks.} We visualize the failure cases observed in four real-world experiments, with key error frames during execution highlighted using red arrows.}
  \label{fig:failure_tasks}
\end{figure*}

\subsection{Further Action Distribution Analysis}
\label{apsec:B4}
We provide further visualization of the action distribution to verify the robustness of our method. Unlike the previous experiment where the green bottle was fixed, we standardize the initial position of the \textbf{red bottle} within the \textit{dual bottles pick hard} task. To provide a more comprehensive analysis of the manipulation strategy, we record the end-effector poses of \textbf{the left arm} across 40 successful trajectories for each model configuration.

We maintain the same comparison settings: (a) SigLIP feature; (b) raw VDM feature; and (c) +Sobel+FFT feature. As illustrated in Figure~\ref{fig:action_distribution2}, the base policy (a) exhibits a \textbf{narrowly concentrated} distribution, indicating a tendency to overfit to a specific subset of expert behaviors. Similarly, the inclusion of unprocessed VDM features (b) fails to significantly expand the diversity of the learned trajectories, resulting in a relatively restricted workspace usage.

In contrast, Video2Act (c), utilizing the refined +Sobel+FFT features, generates a significantly \textbf{broader and more diverse} action distribution. These results further confirm that our method effectively leverages spatio-motional cues to learn robust, multi-modal manipulation strategies that generalize well across different object configurations, rather than collapsing into a single behavioral mode.

\section{Additional Visualizations}
\label{apsec:C}

\subsection{Real-World Grad-CAM Visualizations}
\label{apsec:C1}

We further visualize feature-level Grad-CAM~\cite{selvaraju2017grad} on real-world scenarios and compare DINOv2~\cite{oquab2023dinov2}, SigLIP~\cite{zhai2023siglip}, and our VDM-based representation. As shown in Figure~\ref{fig:real_gradcam}, across both the head camera and wrist camera views, our method exhibits noticeably more stable and coherent activations over time, whereas DINOv2 and SigLIP often produce inconsistent or scattered focus regions. In a few individual frames—such as in the \textit{close laptop} task from the head view or the \textit{pick dual flowers} task from the wrist view—DINOv2 can occasionally localize the target object well. However, in most frames, its attention drifts to irrelevant regions, leading to unstable and unreliable activation patterns. In contrast, VDM representation consistently attends to the manipulated objects and their immediate surroundings, aligning more closely with task-relevant areas. A remaining limitation is that VDM localization becomes less precise when the object and background share nearly identical colors, such as in the fourth frame of the \textit{close laptop} task where both appear white.

\subsection{Simulation Qualitative Results}
\label{apsec:C2}
Figure~\ref{fig:simulation1} and Figure~\ref{fig:simulation2} presents keyframe visualizations of Video2Act executing twelve distinct tasks within the RoboTwin simulation environment. These sequences demonstrate how our dual-system framework effectively translates high-level reasoning into precise, temporally coherent actions.

\noindent\textbf{Bimanual Coordination.} In the \textit{block handover} task (Row 1), Video2Act exhibits precise inter-arm coordination. The left arm stably grasps the object and synchronizes with the right arm for a smooth transfer, validating the effectiveness of our motion-aware features (FFT) in modeling temporal dependencies between two end-effectors. Similarly, in \textit{dual bottles pick easy} (Row 3), the model demonstrates the ability to control both arms simultaneously to grasp upright objects. Crucially, in the \textit{dual bottles pick hard} task (Row 4), where bottles are initialized in random fallen poses, the robot accurately perceives the complex 6D poses and adjusts the gripper orientation for a successful grasp. This highlights that our spatial filtering (Sobel) effectively captures fine-grained structural cues even under significant pose variations.

\noindent\textbf{Robustness in Clutter and Variation.} In the \textit{pick apple messy} task (Row 6), the robot successfully identifies and grasps the target apple amidst distinct distractor objects (e.g., banana, brush), demonstrating that the VDM-based System 2 effectively filters out task-irrelevant visual biases. In \textit{container place} (Row 2) and \textit{empty cup place} (Row 5), the model shows robust spatial reasoning, accurately aligning the grasped object with the target placement zone (plate or coaster) despite variations in object appearance and initial positions.

These qualitative results confirm that Video2Act does not merely memorize trajectories but learns a generalized representation of what to manipulate and how to move, enabling stable execution across diverse dynamic scenarios.

\subsection{Real-World Qualitative Results}
\label{apsec:C3}
Figure~\ref{fig:detailed_tasks} visualizes the execution of six diverse real-world manipulation tasks by Video2Act.These qualitative results highlight three key capabilities of our system:

\noindent\textbf{Dynamic Bimanual Coordination.} In the \textit{handover cucumber} task (Row 5), the system demonstrates \textbf{precise temporal} synchronization. The left arm (giver) and right arm (receiver) align their velocities perfectly during the transfer phase, preventing the object from falling. This verifies that our FFT-based motion extraction effectively captures the inter-arm temporal dependencies required for dynamic handover. Similarly, in the \textit{pick dual carrots} task (Row 3), Video2Act exhibits robust sequential planning. It coordinates the two arms to transport the carrots one after another, maintaining continuous spatial awareness to execute the second grasp accurately without causing interference or collisions with the first arm's trajectory.

\noindent\textbf{Fine-Grained Geometric Reasoning.} The \textit{push triangle} task (Row 6) requires precise spatial reasoning to align a geometric shape with a target slot. Video2Act successfully perceives the orientation of the triangle and plans a pushing trajectory that completes the square pattern, validating the benefit of our Spatial Filtering Operators in capturing object boundaries. Furthermore, in the \textit{pick dual flowers} task (Row 4), the model accurately locates and grasps thin flower stems, which poses a significant challenge for traditional encoders. This demonstrates the superior fine-grained perception of our VDM-based representations during the sequential insertion process.

\noindent\textbf{Articulated Object and Trajectory Modeling.} In the \textit{close laptop} task (Row 2), the robot exhibits an understanding of the articulated object's constraints, generating a smooth, circular trajectory that follows the natural mechanics of the laptop hinge without applying excessive force. Finally, the \textit{pick and place carrot} task (Row 1) confirms the baseline stability of our method in standard pick-and-place scenarios, showing robust grasping and precise release placement into the plate.

\section{Additional Real-World Evaluations}
\label{apsec:D}

\noindent\textbf{Edge cases under lighting variation.}
Following the suggestion, we conduct additional real-robot experiments on the \textit{handover cucumber} task under varying light conditions. We report the performance of Raw, +Sobel, and +Sobel+FFT across four illumination levels in Figure~\ref{fig:lightning_edge}.
The results show that low-light conditions degrade the performance of both Raw and +Sobel, while our proposed combined filter remains relatively robust. We will include more extreme edge-case evaluations in the revised version.

\noindent\textbf{Zero-shot evaluation.}
Following the suggestion, we further evaluate the model on a pick-and-place task involving a Labubu toy, a recently released object not seen during VDM pre-training, achieving an 80\% success rate. In addition, unseen object variations are also shown in Fig.~7 of the main paper.

\begin{table*}[t]
\centering
\caption{\textbf{Generalization performance on the RoboTwin 2.0 benchmark} under three visual conditions: \textit{Clean}, \textit{Background}, and \textit{Light}. All numbers denote success rates.}
\label{tab:generalization_performance}
\resizebox{\textwidth}{!}{
\begin{tabular}{lcccccccccccc}
\toprule
\multirow{2}{*}{\textbf{Task}} 
& \multicolumn{3}{c}{\textbf{RDT}}
& \multicolumn{3}{c}{\textbf{$\pi_0$}}
& \multicolumn{3}{c}{\textbf{$\pi_{0.5}$}}
& \multicolumn{3}{c}{\textbf{Video2Act}} \\
\cmidrule(lr){2-4} \cmidrule(lr){5-7} \cmidrule(lr){8-10} \cmidrule(lr){11-13}
& \cellcolor{condA}\textbf{Clean}
& \cellcolor{condB}\textbf{Background}
& \cellcolor{condC}\textbf{Light}
& \cellcolor{condA}\textbf{Clean}
& \cellcolor{condB}\textbf{Background}
& \cellcolor{condC}\textbf{Light}
& \cellcolor{condA}\textbf{Clean}
& \cellcolor{condB}\textbf{Background}
& \cellcolor{condC}\textbf{Light}
& \cellcolor{condA}\textbf{Clean}
& \cellcolor{condB}\textbf{Background}
& \cellcolor{condC}\textbf{Light} \\
\midrule
Beat Block   & 68.7 & 4.0  & 4.0  & 73.3 & 22.0 & 16.0 & 71.3 & 22.0 & 24.0 & 76.0 & 56.0 & 72.0 \\
Click Bell   & 72.3 & 12.0 & 0.0  & 20.3 & 2.0  & 0.0  & 35.7 & 6.0  & 8.0  & 85.3 & 26.0 & 64.0 \\
Hanging Mug  & 24.3 & 0.0  & 0.0  & 16.3 & 4.0  & 0.0  & 14.0 & 4.0  & 0.0  & 36.7 & 14.0 & 26.0 \\
Place Object & 41.3 & 4.0  & 2.0  & 52.3 & 6.0  & 4.0  & 50.0 & 18.0 & 32.0 & 36.0 & 46.0 & 32.0 \\
Place Shoe   & 36.3 & 2.0  & 2.0  & 62.0 & 6.0  & 10.0 & 58.0 & 18.0 & 24.0 & 50.0 & 20.0 & 34.0 \\
Turn Switch  & 33.0 & 26.0 & 24.0 & 35.0 & 20.0 & 30.0 & 27.7 & 38.0 & 30.0 & 40.7 & 34.0 & 20.0 \\
\midrule
\rowcolor{avggray}
\textbf{Average}
& 46.0 & 8.0 & 5.3
& 43.2 & 10.0 & 10.0
& 42.8 & 17.7 & 19.7
& 54.1 & 32.7 & 43.0 \\
\bottomrule
\end{tabular}
}
\end{table*}

\section{Failure Analysis}
\label{apsec:E}
% Through real-world experiments on the Agilex platform, we observe four specific failure cases encountered by our model.

% \begin{enumerate}
%     \item The first case, observed in the \textit{pick dual flowers} task, reflects a failure in \textbf{manipulation position}. Offsets occur in both the flower picking and flower arranging processes, revealing the difficulty in predicting the position of thin objects.
%     \item In the \textit{pick dual carrots} task, the second case is related to \textbf{incorrect manipulation depth}. The grasp operation is executed before the gripper arrives at the carrot, resulting in failure.
%     \item A \textbf{bimanual dissonance} error could appear in the \textit{handover cucumber} task, which causes the third failure case. The left gripper loosens too fast to allow the right gripper to handle the cucumber.
%     \item The fourth case presents a \textbf{precise operation disability} of the model. While \textit{push triangle}, the gripper tends to move further than needed and break the original shape, failing to build a square in the end.
% \end{enumerate}

Through real-world experiments on the Agilex platform, we categorize the observed failure modes into four distinct types, as visualized in Figure \ref{fig:failure_tasks}:

\begin{enumerate}
    \item The first case illustrates a failure driven by \textbf{common errors} during the \textit{pick dual flowers} task. The extremely thin geometry of the flower stems leads to a minor positional offset during the initial grasp. This slight deviation accumulates throughout the trajectory, causing the robot to drift from the optimal path and ultimately resulting in a misalignment failure when attempting to insert the flower into the vase.
    
    \item The second case, observed in the \textit{pick dual carrots} task, involves \textbf{manipulation height errors} regarding the manipulation height ($z$-axis). The gripper initiates the closing action before reaching the optimal grasping depth for the carrot, resulting in a ``grasping air'' failure.
    
    \item The third case reveals a \textbf{grasp slip} during the \textit{handover cucumber} task. The left gripper (giver) releases the object prematurely before the right gripper (receiver) has established a stable hold, causing the cucumber to fall. This reflects a dissonance in the temporal logic between the two arms.
    
    \item The fourth case presents a limitation in \textbf{minor hardware inaccuracies} during the \textit{push triangle} task. The gripper tends to move further than needed (overshooting), breaking the original shape and failing to build a square in the end. This indicates a deficiency in the model's fine-grained control regarding motion termination.
\end{enumerate}

\noindent To alleviate these limitations, we plan to scale up the collection of high-quality demonstrations and introduce rigorous constraints during training, thereby enhancing robustness in physical environments. Moreover, enabling our System 2 to autonomously detect and rectify erroneous actions will be a key direction for future work.

\section{Additional Quantitative Generalization Experiments}
\label{apsec:F}
To test the generalization ability of the model, we added generalization tests for background randomization and random light in Robotwin 2.0. Background randomization refers to testing the model's adaptability in different background environments, while random light simulates varying lighting conditions to evaluate the model's performance under changing environments. We compared several state-of-the-art methods, including RDT, $\pi_0$, $\pi_{0.5}$, and our method. For each scene, we tested 50 times with different seeds.

As shown in Table~\ref{tab:generalization_performance}, Video2Act demonstrates the highest level of robustness, with the smallest performance degradation under background and lighting variations—specifically drops of 21.4\% (from 54.1\% to 32.7\%) and 11.1\% (from 54.1\% to 43.0\%), respectively. In comparison, $\pi_{0.5}$ shows more significant declines of 25.1\% (42.8\% to 17.7\%) and 23.1\% (42.8\% to 19.7\%). Meanwhile, RDT and $\pi_{0}$ exhibit poor generalization capabilities, with their performance plummeting by 38.0\%, 40.7\% (for RDT), and 33.2\%, 33.2\% (for $\pi_{0}$). These results indicate that the two filters employed in our method effectively mitigate the adverse effects of lighting and background shifts.

\end{document}